\documentclass[conference]{IEEEtran}
\usepackage{times}

\usepackage[numbers]{natbib}
\usepackage{multicol}
\usepackage{amsfonts}
\usepackage[bookmarks=true]{hyperref}
\usepackage{amsmath}
\usepackage{amssymb}  
\usepackage{amsthm}
\usepackage{bm}
\usepackage{mathtools}
\usepackage{color}
\usepackage{graphicx}
\usepackage{booktabs}

\newcommand{\R}{\mathbb{R}}
\newcommand{\nx}{n_x}
\newcommand{\nuu}{n_u}
\newcommand{\Phix}{\mathbf{\Phi}^{\mathrm{x}}}
\newcommand{\Phiu}{\mathbf{\Phi}^{\mathrm{u}}}

\begin{document}

\title{
\looseness-1 \huge Safe Large-Scale Robust Nonlinear MPC in Milliseconds via Reachability-Constrained System Level Synthesis on the GPU}
\author{\authorblockN{Jeffrey Fang and Glen Chou}
\authorblockA{Georgia Institute of Technology, Atlanta, GA 30308\\
Email: \texttt{\{jfang301, chou\}@gatech.edu\vspace{-8pt}}}}

\maketitle

\begin{abstract}
We present GPU-SLS, a GPU-parallelized framework for safe, robust nonlinear model predictive control (MPC) that scales to high-dimensional uncertain robotic systems and long planning horizons. Our method jointly optimizes an inequality-constrained, dynamically-feasible nominal trajectory, a tracking controller, and a closed-loop reachable set under disturbance, all in real-time. To efficiently compute nominal trajectories, we develop a sequential quadratic programming procedure with a novel GPU-accelerated quadratic program (QP) solver that uses parallel associative scans and adaptive caching within an alternating direction method of multipliers (ADMM) framework. The same GPU QP backend is used to optimize robust tracking controllers and closed-loop reachable sets via system level synthesis (SLS), enabling reachability-constrained control in both fixed- and receding-horizon settings. We achieve substantial performance gains, reducing nominal trajectory solve times by 97.7\% relative to state-of-the-art CPU solvers and 71.8\% compared to GPU solvers, while accelerating SLS-based control and reachability by 237$\times$. Despite large problem scales, our method achieves 100\% empirical safety, unlike high-dimensional learning-based reachability baselines. We validate our approach on complex nonlinear systems, including whole-body quadrupeds (61D) and humanoids (75D), synthesizing robust control policies online on the GPU in 20 milliseconds on average and scaling to problems with $2 \times 10^5$ decision variables and $8\times 10^4$ constraints. The implementation of our method is available at \url{https://github.com/Jeff300fang/gpu_sls}.
\end{abstract}

\IEEEpeerreviewmaketitle

\vspace{-3pt}
\section{Introduction}
\vspace{-2pt}

\looseness-1Safe real-time control is essential for long-duration robotic operation. This requires methods to both \textit{synthesize} trajectories and controllers that satisfy safety and task constraints, and \textit{verify} that the resulting closed-loop behavior is robust to \textit{disturbances}. Model-based \textit{trajectory optimization} methods such as nonlinear model predictive control (NMPC) \cite{rawlings2017nmpc} address synthesis, while \textit{reachability analysis} \cite{althoff2021set}, which overapproximates the set of closed-loop reachable states, enables verification.

\looseness-1Despite their success, constrained trajectory optimization and reachability methods struggle to scale to large problems. For legged robots ($>$60 states), NMPC is often too slow for real-time, while nonlinear reachability becomes intractable beyond $\approx$10 states \cite{bansal2017hamilton, landry2018reach, majumdar2017funnel}. Data-driven reachability scales better \cite{bansal2021deepreach, jiang2016using, dawson2022safe}, but the resulting estimates are often inaccurate and compromise safety.
System-level synthesis (SLS) \cite{goulart2006optimization, DBLP:journals/arc/AndersonDLM19} is a scalable alternative, enabling reachability analysis and the co-design of robust control policies \cite{leeman2024fast, leeman2025robust}. However, existing SLS methods are too slow for real-time control of high-dimensional robotic systems. 
While GPU acceleration can enable real-time NMPC \cite{amatucci2025primal, sarkka2022temporal, adabag2024mpcgpu}, existing methods largely ignore inequality constraints and reachability, and therefore cannot guarantee safety under uncertainty.
\begin{figure}[t]
  \centering
  \includegraphics[width=\linewidth]{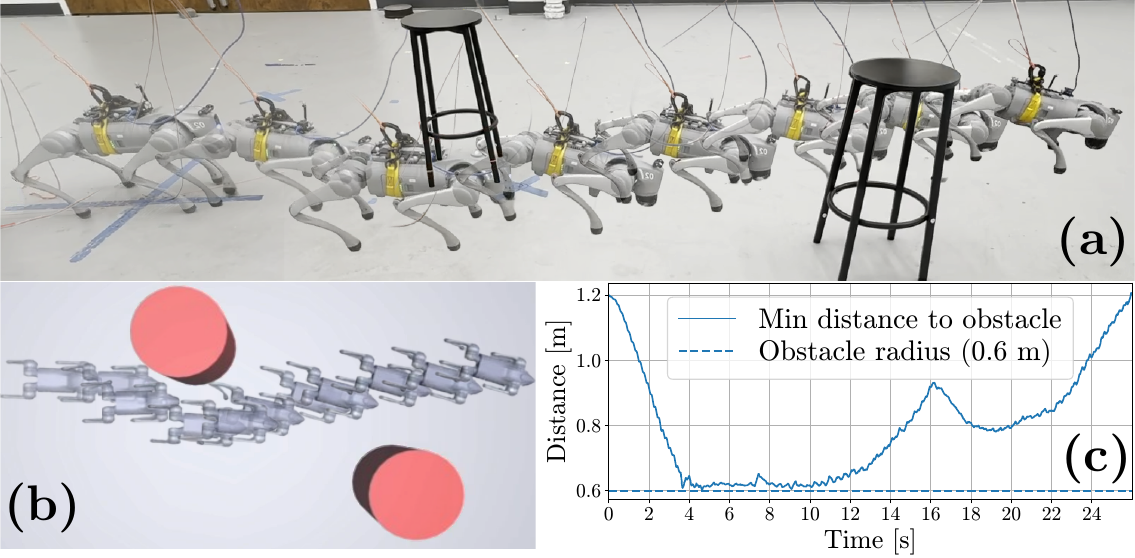}\vspace{-5pt}
  \caption{\textbf{(a)}: GPU-accelerated nonlinear constrained whole-body control (61 states, 12 controls) executed on hardware with a Unitree Go2 EDU quadruped, navigating through an obstacle field in real-time at 50 Hz. \textbf{(b)}: Overhead view of simulated positions reconstructed from hardware-experiment data showing successful navigation around the obstacles. \textbf{(c)}: Graph illustrating minimum distance to any obstacle, demonstrating zero obstacle collisions.}\vspace{-20pt}
  \label{fig:combined_quad}
\end{figure}

\looseness-1To close these gaps, we propose a GPU-parallelized method for robust NMPC ensuring constraint satisfaction under disturbance. \textit{Our key insight is that local dynamics linearizations reduce constrained trajectory planning and reachability analysis to GPU-friendly matrix operations, enabling robust and rapid large-scale robotic control.} We use the alternating direction method of multipliers (ADMM) and the temporal structure of optimal control problems to jointly optimize a constrained nominal trajectory and a robust tracking controller. We achieve \textit{computational efficiency} by caching ADMM iteration-invariant computations and applying logarithmic-depth horizon splitting via parallel associative scans. We ensure \textit{robustness} by computing forward reachable sets of the optimized closed-loop dynamics using SLS on GPU, with an analogous logarithmic-depth parallelization, yielding margins that are used to tighten constraints in the nominal trajectory planning. For a horizon of length $N$ with state and control dimensions $\nx$ and $\nuu$, we achieve a runtime of $\mathcal{O}(\log N\log^2 \nx + \log^2 \nuu)$, improving upon the prior state-of-the-art $\mathcal{O}(N(\nx^3+\nuu^3))$ \cite{leeman2024fast}, and enabling real-time, large-scale robotic control. Our contributions are:
\begin{itemize}
    \item \looseness-1An ADMM-based, GPU-accelerated inequality-constrained linear quadratic regulator (LQR) solver leveraging factorization caching and parallel associative scans.
    \item A real-time NMPC method embedding the LQR solver in a sequential quadratic programming (SQP) loop, enabling nonlinear inequality-constrained trajectory optimization at a cost of $\mathcal{O}(\log N \log^2 \nx + \log^2 \nuu)$ per SQP iteration.
    \item A unified method for real-time reachability, trajectory optimization, and disturbance-feedback synthesis via SLS, with per-iteration complexity $\mathcal{O}(\log N\log^2 \nx + \log^2 \nuu)$.
    \item Evaluation on large-scale problems, including humanoid (75D) control and hardware validation of whole-body quadruped (61D) control, with up to $3\times 10^3$  horizon, $2\times 10^5$ decision variables, and $8 \times 10^4$ constraints, surpassing baselines in safety rate and solve speed.
\end{itemize}

\section{Related Work}

\subsection{Robust Control, Reachability Analysis, and SLS}

Safety verification is commonly performed via reachability analysis~\cite{althoff2021set}, which computes forward invariant sets. Hamilton–Jacobi (HJ) reachability~\cite{bansal2017hamilton} provides strong guarantees but scales poorly due to high-dimensional PDEs, limiting it to low-dimensional systems. Control barrier functions (CBFs)~\cite{ames2019control} offer an alternative, yet synthesizing valid CBFs remains difficult~\cite{dai2022convex}. Data-driven variants scale further~\cite{robey2020learning, saveriano2019learning, so2024train, dawson2022safe} but are often heuristic and may violate safety in practice~\cite{chung2024goal, nath2025formal}; learning-based HJ methods face similar reliability concerns~\cite{bansal2021deepreach}. 
Robust and tube-based MPC~\citep{rawlings2017nmpc, mayne2005robust, kohler2020computationally, langson2004robust, mayne2011tube, 11312909} enforce constraint satisfaction by tightening nominal constraints using reachable sets of the closed-loop system. 
However, directly enforcing closed-loop constraints is typically nonconvex \cite{goulart2006optimization}, leading many methods to rely on conservative overapproximations, often by fixing a feedback policy and computing invariant tubes around nominal trajectories \cite{kousik2020bridging} via sums-of-squares~\cite{singh2018robust, majumdar2017funnel}, HJ reachability~\cite{herbert2017fastrack}, and contraction~\cite{chou2021model, knuth2022statistical, chou2022safe, singh2023robust, knuth2021planning}. While effective, this can be overly conservative. 
In contrast, SLS, also known as disturbance-feedback MPC~\citep{goulart2006optimization, anderson2019system}, offers a convex parameterization of closed-loop responses for linear time-varying (LTV) systems, enabling robust constraint satisfaction under disturbance \citep{chen2024robust, bartos2025stochastic}. Recent work extends SLS to nonlinear dynamics and constraints 
\citep{leeman2025robust, zhan2025robustly}. We build on the SLS framework and leverage GPU parallelism to scale reachability and NMPC to large-scale systems, while maintaining sound reachable-set overapproximations under bounded disturbance.

\subsection{CPU and GPU Parallelization for MPC}
Recent work has focused on accelerating MPC solvers via parallelization and first-order methods like ADMM. TinyMPC~\cite{nguyen2024tinympc} caches factorizations for fast linear solves on embedded platforms, while ReLU-QP~\cite{bishop2024relu} unrolls ADMM iterations as a neural-network-like forward pass for GPU speed. However, these methodologies do not extend to NMPC.
MPCGPU~\cite{adabag2024mpcgpu, adabag2025differentiable, jeon2024cusadi} accelerates NMPC on GPUs but lacks native inequality constraint support, and naive penalty approaches converge slower than our ADMM-based method (Sec. \ref{sec:results}). \cite{jallet2024parallel} reduces horizon complexity by a constant factor through a limited number of parallel trajectory segments, however, does not fully exploit the massive parallelism of GPUs. \cite{pas2025cyqlone} performs cyclic reductions on the KKT system, introducing increased numerical sensitivity, especially under ill-conditioned systems. \cite{jallet2025proxddp} uses an augmented Lagrangian without operator splitting, making convergence sensitive to inexact solves and the penalty parameter~\cite{eckstein2012augmented, eckstein2025two}. ADMM-based OSQP~\cite{stellato2020osqp} uses a direct $LDL^\top$ factorization of the KKT matrix, limiting scalability and parallelization.

Interior-point methods handle inequalities but are hard to parallelize. HPIPM~\cite{frison2020hpipm} exploits dense CPU linear algebra, yet its Riccati recursions create sequential horizon dependencies. Just-In-Time Newton~\cite{iacob2025parallel} adapts interior-point ideas to GPUs using associative scans, but dense floating point computations and multiple scans often underperform optimized CPU baselines. Scan-based trajectory optimization, e.g., Temporal-in-Time~\cite{sarkka2022temporal} and Primal-Dual iLQR~\cite{amatucci2025primal}, enables GPU acceleration, yet constraint handling is limited and robust satisfaction under uncertainty is not addressed. Existing GPU methods mainly target linear or LTI systems, lack nonlinear inequality support, or lack robustness guarantees. Furthermore, second-order methods such as HPIPM are sensitive to ill-conditioning due to their reliance on multiple factorization stages and condensing routines. Limited GPU precision, especially with single-precision arithmetic can degrade stability and convergence behavior. In contrast, first-order methods such as ADMM empirically are more robust to ill-conditioning. Motivated by this, we create a GPU-parallel ADMM approach that can efficiently solve robustly constrained nonlinear MPC with formal guarantees.

\section{Preliminaries and Problem Statement}
\noindent\textbf{Notation:} We use subscripts (e.g. $x_k$) to denote the time index $k$, superscripts $(s)$ for the SQP outer-iteration index and $(t)$ for the ADMM inner-iteration index. We denote the Frobenius norm of a matrix $A \in \mathbb{R}^{m \times n}$ as $\| A \|_{\mathcal{F}}^2 := \text{Trace}(A^\top A)$, $\mathbb{S}_{++}^n$ as the set of symmetric positive definite matrices of size $n \times n$, and $[N]:=\{0,...,N-1\}$ and $[M,N]:=\{M,...,N\}$ for $M,N \in \mathbb{N}$. For a matrix $A \in \mathbb{R}^{n \times p}$ we define the row-wise Euclidean norm $\| A \| _{2,\text{row}} \in \mathbb{R}^{n}$ by $\|A\|_{2, \text{row}} \coloneqq \left[ \|A_{1,:} \|_2, \ldots, \|A_{n,:} \|_2\right]^\top.$

\subsection{Problem Statement}
We consider uncertain discrete-time nonlinear dynamics
\begin{equation}
    \label{eq:nmpc_dynamics}
    x_{k+1} = f(x_k, u_k) + E(x_k) w_k,
\end{equation} 
where $x_k \in \R^{\nx}$ denotes the system state at time step $k$, $u_k \in \R^{\nuu}$ denotes the control input, $f: \mathbb{R}^{n_x} \times \mathbb{R}^{n_u} \rightarrow \mathbb{R}^{n_x}$ the dynamics function,
$E: \mathbb{R}^{n_x} \rightarrow \mathbb{R}^{n_x \times n_x}$ the disturbance scaling function, and $w_k \in \mathcal{E}_{n_x} \coloneqq \{ w \in \mathbb{R}^{n_x}, \|w\|_2 \leq 1 \}$ the disturbance, normalized to be contained in a unit ball.

We consider the problem of designing an optimal controller $\pi(\cdot)$ for the following \textit{robust} nonlinear control problem:
\begin{subequations} \label{eq:robust_nocp}
    \begin{align}
        \min_{\pi(\cdot)} \quad &J\left(\bar{x}, \pi(\cdot)\right) \\
        \text{s.t.} \quad & x_{k+1} = f(x_k, u_k) + E(x_k)w_k, \quad \forall k \in [N],
        \\
                          & x_0 = \bar{x}_0, \\ 
                          & u_k = \pi_k(x_{0:k}), \quad \forall k \in [N], \\
                          & g(x_k, u_k) \leq 0, \quad\forall k \in [N], \quad \forall w_k \in \mathcal{E}_{n_x}, \\
                          & g^f(x_N) \leq 0, \quad \forall w_N \in \mathcal{E}_{n_x},
    \end{align}
\end{subequations}
\looseness-1where $\pi = (\pi_0, \ldots, \pi_N)$ is a sequence of causal control policies, $\bar x_0 \in \mathbb{R}^{\nx}$ is the initial state, $g: \mathbb{R}^{n_x} \times \mathbb{R}^{n_u} \rightarrow \mathbb{R}^{n_c}$ denotes the stagewise state-input constraints, and $g^f: \mathbb{R}^{n_x} \rightarrow \mathbb{R}^{n_f}$ the terminal constraint. As \eqref{eq:robust_nocp} is an intractable infinite-dimensional policy optimization, approximate methods typically decompose it by solving for (A) a nominal state-input trajectory and (B) a feedback controller around this nominal solution. Our solution uses the formalism of NMPC for (A) (Sec. \ref{sec:nmpc}) and system level synthesis (SLS) for (B) (Sec. \ref{sec:sls}).

\subsection{Nonlinear Model Predictive Control (NMPC)}\label{sec:nmpc}
NMPC is a strategy that repeatedly solves the following optimal control problem given an initial state $\bar{x}_0 \in \mathbb{R}^{\nx}$
\begin{subequations} \label{eq:nmpc_problem}
    \begin{align}
        \min_{\substack{
        X,\, U
        }} \quad
        & J(X,U) := \ell_f(x_N) + \textstyle\sum_{k=0}^{N-1} \ell(x_k, u_k) \label{eq:nmpc_problem_obj} \\
        \text{s.t.} \quad 
        & x_{k+1} = f(x_k, u_k), \quad\quad  \forall k \in [N],\label{eq:dynamics} \\
        & x_0 = \bar{x}_0, \label{eq:nmpc_init} \\
        & g(x_k, u_k) \leq 0, \hspace{4pt}\forall k \in [N], \quad g^f(x_N) \leq 0, \label{eq:constraints}
    \end{align}
\end{subequations}
\looseness-1yielding an \textit{open-loop} \textit{nominal} state trajectory $X := \{x_k\}_{k=0}^N$ and control sequence $U := \{u_k\}_{k=0}^{N-1}$ over a horizon of length $N$ that minimizes a nonlinear cost function $J: \mathbb{R}^{\nx (N+1)} \times \mathbb{R}^{\nuu N}$ 
with running $\ell: \mathbb{R}^{\nx} \times \mathbb{R}^{\nuu} \rightarrow \mathbb{R}$ and terminal costs $\ell_f: \mathbb{R}^{\nx} \rightarrow \mathbb{R}$, subject to the \textit{nominal} dynamics \eqref{eq:dynamics}, and the constraints \eqref{eq:constraints}.
It is commonly solved using sequential quadratic programming (SQP) \cite{nocedal2006numerical}, which linearizes the nonlinear dynamics and constraints and forms a quadratic approximation of the Lagrangian around a nominal trajectory.
At each SQP iteration, we form the associated Lagrangian
\begin{equation} \label{eq:lagrangian}
    \begin{aligned}
      \mathcal{L}(X,U, \mu, \gamma, \nu) &= J(X,U) + \mu_0^\top (\bar{x}_0 - x_0) \\ 
      & \quad + \textstyle\sum^{N-1}_{k=0} \mu_{k+1}^\top \left(f(x_k, u_k) - x_{k+1}\right) \\
      & \quad + \textstyle\sum^{N-1}_{k=0} \gamma_k^\top g(x_k, u_k) + \nu^\top g^f(x_N),
    \end{aligned}
\end{equation}
where $\mu_k \in \mathbb{R}^{n_x}$ are the Lagrange multipliers for the dynamics constraints, $\gamma_k \in \mathbb{R}_{\ge 0}^{n_c}$ for the stage inequality constraints, and $\nu \in \mathbb{R}_{\ge 0}^{n_f}$ for the terminal inequality constraint.
Linearizing the dynamics and constraints around the current nominal trajectory $\xi^{(s)}:=(x^{(s)}, u^{(s)})$ and forming a quadratic approximation of the Lagrangian \eqref{eq:lagrangian} yields the structured LTV-QP \eqref{eq:sqp_qp}:
\begin{subequations} \label{eq:sqp_qp}
    \begin{align}
        \hspace{-10pt}\min_{\delta x, \delta u} \quad & J_\text{QP}(\delta x, \delta u):= \hspace{-3pt}\sum_{k=0}^{N-1}
        \frac{1}{2} \begin{bmatrix}
        \delta x_k \\
        \delta u_k
        \end{bmatrix}^\top
        \hspace{-4pt}\begin{bmatrix}
        Q_k & S_k^\top \\
        S_k & R_k
        \end{bmatrix}\hspace{-3pt}
        \begin{bmatrix}
        \delta x_k \\
        \delta u_k
        \end{bmatrix}\hspace{-10pt} \nonumber\\
        & + 
        \begin{bmatrix}
        q_k \\
        r_k
        \end{bmatrix}^\top
        \begin{bmatrix}
        \delta x_k \\
        \delta u_k
        \end{bmatrix}
        + \frac{1}{2} \delta x_{N}^\top Q_N \delta x_N+ q^\top_N \delta x_N \\
        \text{s.t.} \quad
        & \delta x_{k+1} = A_k \delta x_k + B_k \delta u_k + b_k,\quad \forall k\in [N],\label{eq:linear_dynamics} \\
        & \delta x_0 = \bar{x}_0 - x^{(s)}_0, \\
        & C_k \delta x_k + D_k \delta u_k \leq f_k,\hspace{1pt}\quad \forall k\in [N], \label{eq:lin_constraints_stage} \\ 
        & C_N \delta x_N \leq f_N, \label{eq:lin_constraints_term}
    \end{align}
\end{subequations}
where
\begin{subequations}
    \begin{align}
        & \delta x_k = x_k - x_k^{(s)} \qquad \delta u_k = u_k - u_k^{(s)} \\
        & A_k = \nabla_x f \mid_{\xi_k^{(s)}}, \qquad B_k = \nabla_u f \mid_{\xi_k^{(s)}}, \\
        & C_k = \nabla_x g \mid_{\xi_k^{(s)}}, \qquad D_k = \nabla_u g \mid_{\xi_k^{(s)}}, \\
        & C_N = \nabla_x g_f \mid_{x_N^{(s)}} ,\\
        & f_k = -g(x_k^{(s)}, u_k^{(s)}), \quad f_N = -g_f(x_N^{(s)}).
    \end{align}
\end{subequations}
The quadratic terms are approximated by
\begin{subequations}
    \begin{align}
        & \hspace{-2pt} \begin{bmatrix}
        Q_k & S_k^\top \\
        S_k & R_k
        \end{bmatrix}
        = \nabla^2_{(x_k, u_k)}\mathcal{L}_k \mid_{\xi_k^{(s)}}, \quad Q_N = \nabla^2_{x_k} \mathcal{L}_{N} \mid_{x_N^{(s)}}, \\
        & q_k = \nabla_{x_k} \mathcal{L}_k|_{\xi_k^{(s)}} \qquad r_k = \nabla_{u_k} \mathcal{L}_k|_{\xi_k^{(s)}} \\ 
        & q_N = \nabla_{x_N} \mathcal{L}_N|_{x_N^{(s)}}.
    \end{align}\hspace*{-8pt}
\end{subequations} 
The solution of the LTV-QP \eqref{eq:sqp_qp} yields search direction $\delta \xi:=(\delta x, \delta u)$, which updates the nominal iterate as $x^{(s+1)} = x^{(s)} + \alpha \delta x$ and $u^{(s+1)} = u^{(s)} + \alpha \delta u$, 
with step size $\alpha \in (0, 1]$.  This reduces NMPC to a sequence of constrained optimal control problems with linear time-varying (LTV) dynamics, posed as quadratic programs (QPs), whose solutions iteratively update the nominal trajectories \cite{rawlings2017nmpc}.

\subsection{Alternating Direction Method of Multipliers (ADMM)}\label{sec:admm}
The ADMM \cite{boyd2011distributed} is a first-order method to efficiently solve constrained QPs. We consider a generic QP of the form
\begin{equation}
    \label{eq:generic_qp}
    \begin{aligned}
        \min_{x \in \mathcal{C}} \quad &
        f(x)
    \end{aligned}
\end{equation}
\looseness-1where $f: \mathbb{R}^n \rightarrow \mathbb{R}$ is convex and $\mathcal{C} \subseteq \mathbb{R}^n$ is a convex set. 
ADMM absorbs the set constraints $x \in \mathcal{C}$ into the objective via a split variable $z$ and an indicator function, reformulating \eqref{eq:generic_qp} into
\begin{subequations}
    \begin{align}
        &\min_{x,z} \quad
            f(x) + I_\mathcal{C}(z) \qquad \text{s.t.} \quad x = z, \label{eq:admm_split_constraint} \\
        & I_\mathcal{C}(z) = \begin{cases}
            0 & z \in \mathcal C \\
            + \infty & \mathrm{otherwise},
        \end{cases} 
    \end{align}
\end{subequations}
and solves it using an augmented Lagrangian
\begin{equation}
    \label{eq:aug_lagrangian}
    \mathcal{L}_{A}(x,z, \lambda) := f(x) + I_\mathcal{C}(z) + \lambda^\top(x-z) + \textstyle\frac{\rho}{2} \| x - z\|_2^2,
\end{equation}
where $\lambda \in \mathbb{R}^{n}$ is the Lagrange multiplier and $\rho \in \mathbb{R}_{>0}$ is a scalar penalty parameter.
ADMM proceeds by alternating between 1) minimizing \eqref{eq:aug_lagrangian}
with respect to the primal variables $x$ and $z$, and 2) performing a gradient ascent step on the dual variable $\lambda$, i.e., at iteration $t$, ADMM performs the updates
\begin{align}
    x^{(t + 1)} := & \arg\min_{x} \mathcal{L}_A (x, z^{(t)}, \lambda^{(t)}),
    \label{eq:admm_x_update}
    \\
    z^{(t+1)} := & \arg\min_{z} \mathcal{L}_A (x^{(t+1)}, z, \lambda^{(t)}),
    \label{eq:admm_z_update}
    \\
    \lambda^{(t+1)} := & \lambda^{(t)} + \rho (x^{(t+1)} - z^{(t+1)}).
    \label{eq:admm_lam_update}
\end{align}
In ADMM formulations tailored for QPs, \eqref{eq:admm_x_update} corresponds to the solution of an unconstrained QP, \eqref{eq:admm_z_update} reduces to a projection onto $\mathcal{C}$, and the dual variable $\lambda$ is updated via a simple ascent step \eqref{eq:admm_lam_update}. By iterating over updates \eqref{eq:admm_x_update}-\eqref{eq:admm_lam_update}, ADMM is guaranteed to converge to an optimal solution of \eqref{eq:generic_qp} 
\cite{boyd2011distributed}.

\subsection{System Level Synthesis}\label{sec:sls}
NMPC alone does not robustly guarantee safety under disturbance. To do so, we unify NMPC with SLS, which optimizes over causal \textit{disturbance feedback} controllers
\begin{equation}\label{eq:disturbance_feedback}
u_k = v_k + \textstyle\sum_{j=0}^{k-1} \mathbf{\Phi}_{k,j}^u w_j,
\end{equation}
\looseness-1with nominal control $v_k \in \R^{\nuu}$, i.e., we assign a distinct disturbance feedback matrix $\mathbf{\Phi}_{k,j}^u\in \R^{\nuu\times \nx}$ for each disturbance $w_j$ and control $u_k$ for $k>j$. For uncertain LTV dynamics
\begin{equation}\label{eq:dynamics_ltv}
    x_{k+1} = A_kx_k + B_ku_k+E_kw_k,
\end{equation}
it can be shown using standard algebraic manipulations \cite{anderson2019system} that the resulting closed-loop state sequence can be expressed as
\begin{equation}\label{eq:SLS_state}
x_k= z_k   + \textstyle\sum_{j=0}^{k-1} \mathbf{\Phi}_{k,j}^x w_j,~ z_0 = \bar x_0,
\end{equation}
where $z_k \in \mathbb{R}^{\nx}$ is the nominal state and \mbox{$\mathbf{\Phi}_{k,j}^x\in \R^{\nx\times \nx}$} captures the influence of disturbance $w_j$ on state $x_k$.
Starting with $\mathbf{\Phi}_{j+1,j}^x = E_j$, SLS propagates the disturbance via 
\begin{equation}
\label{eq:tube_propagation}
\begin{aligned}
\mathbf{\Phi}_{k+1,j}^x = A_k \mathbf{\Phi}_{k,j}^x + B_k\mathbf{\Phi}_{k,j}^u,\\
\end{aligned}
\end{equation}
for all $j \in [N]$ and $k \in [j+1,N-1]$. Since \eqref{eq:SLS_state} and \eqref{eq:disturbance_feedback} describe the \textit{true} closed-loop trajectory under a disturbance sequence, enforcing constraints over a worst-case disturbance set ensures robustness. 
Using this idea, SLS can be extended to nonlinear systems by planning a nominal trajectory and modeling tracking errors as an LTV system \eqref{eq:dynamics_ltv}, 
where $E_k$ can be chosen to bound linearization error \cite{zhan2025robustly, leeman2025robust}. We note that we do not formally consider linearization error, however, such bounds can be formally incorporated following \cite{leeman2025robust}, albeit at the cost of increased conservativeness. This yields an approximate solution to the robust NMPC problem \eqref{eq:robust_nocp}:
\begin{subequations}\label{eq:fastsls_problem}
    \begin{align}
        \hspace{-13pt}\min_{\substack{
        X,\, U, \mathbf{\Phi}
        }} \ \
        & J(X,U) + \tilde{H}_0(\mathbf{\Phi}) \label{eq:sls_problem_obj} \\
        \text{s.t.}\quad \
        & x_{k+1} = f(x_k, u_k), \quad \forall k \in [N],\quad x_0 = \bar{x}_0,\label{eq:sls_problem_dyn} \\ 
        & \mathbf{\Phi}^\text{x}_{k+1, j} = A^{(s)}_k\mathbf{\Phi}^\text{x}_{k, j} + B^{(s)}_k\mathbf{\Phi}^\text{u}_{k,j}, \\
        & \nonumber \quad\quad\forall j \in [N], \hspace{5pt} \forall k \in [j+1, N - 1], \\
        & \mathbf{\Phi}^\text{x}_{j+1,j} = E(x_j), \\
        & g(x_k, u_k) + h_{k}(\mathbf{\Phi}) \leq 0, \qquad \forall k \in [N],\hspace{-5pt}\label{eq:rmpc_stage_tightening}\\
        & g^f(x_N) + h^f(\mathbf{\Phi}) \leq 0\label{eq:rmpc_term_tightening},
    \end{align}
\end{subequations}
where $A^{(s)}_k, B^{(s)}_k$ are the linearized dynamics \eqref{eq:linear_dynamics} at stage $k$. We define the stacked constraint tightenings as 
\begin{subequations}
    \begin{align}
        & h_k(\mathbf{\Phi})
        =
        \textstyle\sum_{j=0}^{k-1}
        \big\Vert
        \big(C^{(s)}_{k}\big)\mathbf{\Phi}^{\mathrm{x}}_{k,j}
        +
        \big(D^{(s)}_{k}\big)\mathbf{\Phi}^{\mathrm{u}}_{k,j}
        \big\Vert_{2,\mathrm{row}} \\
        & h^f(\mathbf{\Phi})
        =
        \textstyle\sum_{j=0}^{N-1}
        \big\Vert
        \big(C^{(s)}_{N}\big)\mathbf{\Phi}^{\mathrm{x}}_{N,j}
        \big\Vert_{2,\mathrm{row}}. 
    \end{align}
\end{subequations}
where $C^{(s)}_{k}, D^{(s)}_{k}, C^{(s)}_{N}$ are the linearized constraints \eqref{eq:lin_constraints_stage}-\eqref{eq:lin_constraints_term}. The vectors $h_k(\mathbf{\Phi})$ and $h^f(\mathbf{\Phi})$ quantify conservative margins induced by the propagation of disturbances through the closed-loop dynamics, and are used to robustly enforce the constraints. We define
\begin{equation}
    \begin{aligned}
        \tilde{H}_0(\mathbf{\Phi}) &=\sum_{j=0}^{N-1}\Big(\sum_{k=j}^{N-1}\!\left(\|\bar Q^\frac{1}{2}\mathbf{\Phi}^{\mathrm{x}}_{k,j}\|_{\mathcal F}^2+\|\bar R^\frac{1}{2}\mathbf{\Phi}^{\mathrm{u}}_{k,j}\|_{\mathcal F}^2\right) \\
        & +\|\bar Q_N^\frac{1}{2}\mathbf{\Phi}^{\mathrm{x}}_{N,j}\|_{\mathcal F}^2 \Big), 
    \end{aligned}
\end{equation}
where  $\mathbf{\Phi}$ collects all $\mathbf{\Phi}^{\text{x}}, \mathbf{\Phi}^\text{u}$ and $\bar{Q} \in \mathbb{S}_{++}^{n_x}, \bar{R} \in \mathbb{S}_{++}^{n_u}, \bar{Q}_N \in \mathbb{S}_{++}^{n_x}$ define a quadratic cost on the system-level responses $\mathbf{\Phi}$.
The state-of-the-art solver for \eqref{eq:fastsls_problem} is FastSLS \cite{leeman2024fast}, which decomposes \eqref{eq:fastsls_problem} into an iterative procedure that alternates between optimizing a (A) nominal trajectory and a (B) robust controller. The nominal trajectory $(X, U)$ is computed by solving a constraint-tightened NMPC,
\begin{subequations} \label{eq:nmpc_problem_tightened}
    \begin{align}
        \hspace{-5pt}\min_{\substack{
        X,\, U
        }} \quad
        & J(X,U) \\
        \text{s.t.} \quad 
        & x_{k+1} = f(x_k, u_k), \hspace{2pt} \forall k \in [N], \qquad x_0 = \bar{x}_0, \\
        & g(x_k, u_k) + h_k(\mathbf{\Phi})\leq 0,\qquad  \forall k \in [N], \\
        & g^f(x_N) + h^{f}(\mathbf{\Phi})\leq 0.
    \end{align}
\end{subequations}
By defining $g^*(x_k, u_k) \coloneqq g(x_k, u_k) + h_k(\mathbf{\Phi})$ and $g^{f^*}(x_N) \coloneqq g^f(x_N) + h^{f}(\mathbf{\Phi})$, \eqref{eq:nmpc_problem_tightened} mirrors the structure of \eqref{eq:robust_nocp} and can be solved using the method described in Sec. \ref{sec:nmpc} and Sec. \ref{sec:fast_qp}.
After the nominal trajectory update, FastSLS solves for the robust controller update to obtain $\mathbf{\Phi}$ using: 
\begin{subequations} \label{eq:fastsls_qp}
    \begin{align}
        \hspace{-3pt}\min_{\mathbf{\Phi} ^ \text{x}, \mathbf{\Phi}^{\text{u}}} \ 
        &\sum_{j=0}^{N-1}\hspace{-3pt} \left(
        \sum_{k=j}^{N-1} \| \mathcal{Q}_{k,j} \mathbf{\Phi}_{k,j} \|^2_{\mathcal{F}} + \|\mathcal{Q}_{N,j} 
         \mathbf{\Phi}^\text{x}_{N,j} \|^2_{\mathcal{F}}\right) \\
        \text{s.t.} \ & \mathbf{\Phi}^{\text{x}}_{k+1,j} = A^{(s)}_k \mathbf{\Phi}^{x}_{k,j} + B_k^{(s)} \mathbf{\Phi}^{\text{u}}_{k,j}, \\
        & \mathbf{\Phi}^{\text{x}}_{j+1,j} = E_j, \ \ \forall j \in [N], \ \ \forall k \in [j+1,N-1],
    \end{align}
\end{subequations}
where $\mathbf{\Phi}_{k,j} := \begin{bmatrix}
    {\Phix_{k,j}}^{\top} & \hspace{-2mm}{\Phiu_{k,j}}^{\top}
\end{bmatrix}^{\top}$. We define the concatenation and block decomposition
\begin{equation}\label{eq:sls_cost_matrices_tau}
    \begin{aligned}
        \hspace{-5pt}& \mathcal{Q}_{k,j} = \left( \text{diag}\left( \sqrt{\tau_{k,j}}\right)
        \begin{bmatrix} C_k^{(s)} & D_k^{(s)} \end{bmatrix},
        \begin{bmatrix}
            \bar{Q}^\frac{1}{2} & 0\\
            0 & \bar{R}^\frac{1}{2}
        \end{bmatrix}
        \right), \\
        \hspace{-5pt}& \mathcal{Q}_{N,j} = \left( \text{diag}\left( \sqrt{\tau_{N,j}}\right) C_N^{(s)}, \bar{Q}_N^{\frac{1}{2}}\right), \\
        & \begin{bmatrix}
            \mathcal{Q}^{\text{x}}_{k,j} & \mathcal{Q}^{\text{xu}}_{k,j} \\
            \mathcal{Q}^{\text{ux}}_{k,j} & \mathcal{Q}^{\text{u}}_{k,j}
        \end{bmatrix}
        \coloneqq \mathcal{Q}^\top_{k,j} \mathcal{Q}_{k,j}.
    \end{aligned}
\end{equation}
where $\tau$ is the Lagrange multiplier that enforces consistency between the nominal optimization and controller synthesis and $\mathcal{Q}^{\text{x}}_{k,j} \in \mathbb{S}^{n_x}_{++}, \mathcal{Q}^{\text{u}}_{k,j} \in \mathbb{S}^{n_u}_{++}, \mathcal{Q}^{\text{xu}}_{k,j} \in \mathbb{R}^{n_x \times n_u}, \mathcal{Q}^{\text{ux}}_{k,j} \in \mathbb{R}^{n_u \times n_x}$. 

\begin{figure*}
    \centering
    \vspace{-12pt}
    \includegraphics[width=\linewidth]{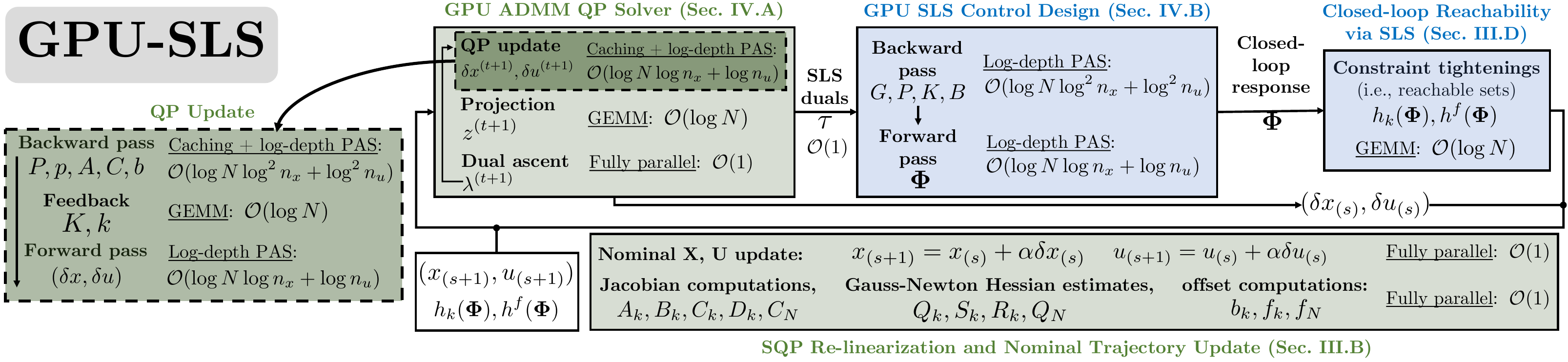}
    \vspace{-20pt}
    \caption{\looseness-1\textbf{Schematic of our method, GPU-SLS.}
At each SQP iteration $(s)$, we compute a nominal trajectory update $(\delta x, \delta u)$ using a GPU-parallelized ADMM-based QP solver that exploits caching and parallel associative scans (PASs) for acceleration. The resulting dual variables are post-processed into SLS-compatible duals $\tau$ (App.~\ref{app:fastsls_duals}), which inform the objective in the SLS controller optimization problem~\eqref{eq:sls_problem_obj}. This optimization is likewise GPU-parallelized via PASs, yielding closed-loop response matrices $\mathbf{\Phi}$ that implicitly define the controller. 
These matrices define constraint tightenings~\eqref{eq:rmpc_stage_tightening} and~\eqref{eq:rmpc_term_tightening}, i.e., closed-loop reachable sets, which are used to tighten the constraints of the subsequent nominal trajectory solve, ensuring robust feasibility under closed-loop control. The next SQP iteration then proceeds after fully parallelizable Jacobian and Hessian updates, along with state, input, and function evaluations. Green denotes nominal trajectory computations; blue denotes feedback controller and reachable set computations. GEMM is a $\mathcal{O}(\log N)$ matrix multiplication routine.
    }\vspace{-15pt}
    \label{fig:schematic}
\end{figure*}

\section{Method}
\looseness-1We present an efficient framework to solve \eqref{eq:robust_nocp} using NMPC and SLS on the GPU. First, Sec. \ref{sec:fast_qp} introduces a GPU accelerated ADMM-based method for efficiently solving the structured LTV-QPs in each SQP iteration, accelerating nominal NMPC. Second, Sec. \ref{sec:method_sls} details our parallelized solution of the robust SLS control synthesis problem, accelerating control design and reachability analysis. Sec. \ref{sec:rti} describes the Real-Time Iteration (RTI) approach used to enable online simultaneous reachability analysis and trajectory optimization.

\subsection{GPU ADMM QP Solver} \label{sec:fast_qp}
We solve LTV-QPs of the form \eqref{eq:sqp_qp} using a GPU-parallelized ADMM QP solver. We introduce the split variable
$z \coloneqq (z_0, \ldots, z_N)$
that tracks the left-hand side of the linearized inequality constraints \eqref{eq:lin_constraints_stage}-\eqref{eq:lin_constraints_term}, i.e., we define
\begin{equation}
    \begin{aligned}
        z_k = C_k \delta x_k + D_k \delta u_k, \quad \forall k \in [N]; \quad z_N = C_N \delta x_N.
    \end{aligned}
\end{equation}
Let $f \coloneqq (f_0, \ldots, f_N)$ denote the stacked constraint offsets. The inequality constraints in \eqref{eq:sqp_qp}, i.e., \eqref{eq:lin_constraints_stage}-\eqref{eq:lin_constraints_term}, can then equivalently be written as $z \leq f$ (elementwise). For compactness, we define $\delta \xi_k:=(\delta x_k, \delta u_k)$ and the linear mapping $G: \mathbb{R}^{\nx(N+1)}\times\mathbb{R}^{\nuu N} \rightarrow \mathbb{R}^{n_cN + n_f}$,
\begin{equation}
    \begin{aligned}
        G(\delta \xi):=G(\delta x, \delta u) = (G_0, \ldots, G_N),\quad \textrm{with}\\
        G_k (\delta \xi_k) \coloneqq C_k \delta x_k + D_k \delta u_k, \quad G_N(\delta x_N) \coloneqq C_N \delta_N.
    \end{aligned}
\end{equation}
The split constraints \eqref{eq:admm_split_constraint} can be then written as $G(\delta x, \delta u) = z$. Define the convex set $\mathcal{C} \coloneqq \{ z \in \mathbb{R}^{n_cN + n_f} : z \le f \}$, we can equivalently form \eqref{eq:sqp_qp} as
\begin{equation} \label{eq:admm_qp}
    \begin{aligned}
        \min_{\delta x, \delta u} \quad & J_{\text{QP}}(\delta x, \delta u) + I_{\mathcal{C}}(z) \\
        \text{s.t.} \quad & \delta x_{k+1} = A_k \delta x_k + B_k \delta u_k + b_k,\quad \forall k \in [N], \\ 
        &  G(\delta x, \delta u) = z. \\
    \end{aligned}
\end{equation}
Now, we define the augmented Lagrangian of \eqref{eq:admm_qp}.
\begin{equation} \label{eq:al_admm}
    \begin{aligned}
        & \mathcal{L}_A(\delta \xi, \lambda, \rho) = J_{\text{QP}}(\delta \xi) + I_\mathcal{C}(z)+ \lambda^T (G(\delta \xi) - z)\\ 
        & \hspace{64pt} + \textstyle\frac{\rho}{2} \|G(\delta \xi) - z \|_2^2 \\
        &\text{s.t.} \hspace{6pt} \delta x_{k+1} = A_k \delta x_k + B_k \delta u_k + b_k, \hspace{5pt} \delta x_0 = \bar{x}_0 - x^{(s)}_0,
    \end{aligned}
\end{equation}
where $\lambda$ is the Lagrange multiplier associated with the consensus constraint and $\rho$ is the penalty parameter. ADMM proceeds by \eqref{eq:admm_x_update}, \eqref{eq:admm_z_update}, and \eqref{eq:admm_lam_update} to solve this optimal control problem.

First, the primal update (equivalent of \eqref{eq:admm_x_update}) reduces to solving an equality-constrained QP of the form
\begin{subequations} \label{eq:primal_update}
    \begin{align}
        \min_{\delta x, \delta u} \quad &\sum_{k=0}^{N-1} \nonumber
        \frac{1}{2} \begin{bmatrix}
        \delta x_k \\
        \delta u_k
        \end{bmatrix}^\top
        \begin{bmatrix}
        \hat{Q}_k & \hat{S}_k^\top \\
        \hat{S}_k & \hat{R}_k
        \end{bmatrix}
        \begin{bmatrix}
        \delta x_k \\
        \delta u_k
        \end{bmatrix}
        + 
        \begin{bmatrix}
        \hat{q}_k \\
        \hat{r}_k
        \end{bmatrix}^\top
        \begin{bmatrix}
        \delta x_k \\
        \delta u_k
        \end{bmatrix} \\
        & + \frac{1}{2} \delta x_{N}^\top \hat{Q}_N \delta x_N
        + \hat{q}^\top_N \delta x_N \\
        \text{s.t.} \quad
        & \delta x_{k+1} = A_k \delta x_k + B_k \delta u_k + b_k,\ \forall k \in [N], \\
        & \delta x_0 = \bar{x}_0 - x^{(s)}_0,
    \end{align}
\end{subequations}
where the augmented costs are defined as
\begin{equation} \label{eq:augmented_matrices}
\hspace*{-0.4em}
\renewcommand{\arraystretch}{1.3}
    \begin{array}{l@{\quad}l}
        \hat{Q}_k = Q_k + \rho C_k^\top C_k, & \hat{q}_k = q_k + C_k^\top(\lambda_k - \rho z_k), \\
        \hat{R}_k = R_k + \rho D_k^\top D_k, & \hat{r}_k = r_k + D_k^\top(\lambda_k - \rho z_k), \\
        \hat{S}_k = S_k + \rho C_k^\top D_k, & \forall k \in [N], \\
        \hat{Q}_N = Q_N + \rho C_N^\top C_N, & \hat{q}_N = q_N + C_N^\top(\lambda_N - \rho z_N).
    \end{array}\hspace{-10pt}
\end{equation}
As in common ADMM practice, we reformulate \eqref{eq:augmented_matrices} by introducing the scaled dual variables $y:=(y_0,\ldots,y_N)$, where $y_k := \lambda_k/\rho$ for all $k \in [N]$ and $y_N :=\lambda_N/\rho$: 
\begin{equation} \label{eq:augmented_linear}
\hspace*{-0.32em}
    \begin{aligned}
        &\hat{q}_k = q_k + \rho C^\top_k(y_k- z_k), \quad \hat{r}_k = r_k + \rho D^\top_k(y_k- z_k), \\
        &\hat{q}_N = q_N + \rho C^\top_N(y_N- z_N).
    \end{aligned}
\end{equation}
\looseness-1Next, the $z$-update \eqref{eq:admm_z_update} becomes a simple linear projection onto the convex set $\mathcal{C}$, after which the dual ascent step \eqref{eq:admm_lam_update} is performed as vector operations:
\begin{subequations}
    \begin{align}
        &z^{(t+1)} = \min\big(G(\delta x^{(t + 1)}, \delta u^{(t + 1)}) + y^{(t)}, f\big), \\
        &\lambda^{(t+1)} = \lambda^{(t)} + \rho \big(G(\delta x^{(t+1)}, \delta u^{(t+1)}) - z^{(t+1)}\big).
    \end{align}
\end{subequations}

\noindent Note that $z$- and $\lambda$-updates are fast and fully component-wise parallelizable across time and constraints. Thus, the dominant computational cost of our ADMM formulation is in the primal update \eqref{eq:primal_update}. To accelerate this structured equality-constrained solve on GPUs, we build on recent GPU-parallel optimal control solvers that exploit temporal parallelism via associative scans \cite{amatucci2025primal}.

\subsubsection{Efficiently Solving \eqref{eq:primal_update} via Associative Scans}
Note that \eqref{eq:primal_update} is a linear quadratic regulator (LQR) problem, which is traditionally solved via Riccati recursions \cite{rawlings2017nmpc}, yielding an optimal solution of the form $(\delta x, \delta u := K \delta x+k)$ \cite{todorov2005generalized}. However, Riccati recursions are solved sequentially across time-steps, leading to a computational bottleneck. To accelerate the solution of \eqref{eq:primal_update}, we use \emph{reverse parallel associative scan} operations to \textit{temporally parallelize} the solution of the primal update \eqref{eq:primal_update}. Given elements $a_1, \ldots, a_N$ and an associative binary operator $\otimes$, a reverse parallel associative scan computes a sequence of suffix reductions $(s_1, s_2, \ldots, s_N) := s_{1:N}$ where
\begin{equation} \label{eq:revere_assoc_scan}
    s_{1:N} := (a_1 \otimes a_2 \otimes \cdots \otimes a_N,\ a_2 \otimes \cdots \otimes a_N,\ \ldots\ ,a_N),
\end{equation}
in $\mathcal{O} (\log N)$ depth on parallel hardware \cite{blelloch2002scans}. A more detailed overview is given in App. \ref{app:parallel_associative_scan}. To exploit this associative scan to solve \eqref{eq:primal_update}, we first define the conditional value function (CVF) of \eqref{eq:primal_update}, building on the result of \cite{amatucci2025primal} and extending it to solve the modified LQR problem given by ADMM \eqref{eq:primal_update}:
\begin{equation} \label{eq:CVF}
    \begin{aligned}
        V_{i \rightarrow j}(x_i, x_j) = \max_\eta &\frac{1}{2} x_i^\top\tilde{P}_{i,j} x_i + \tilde{p}_{i,j}^\top x_i - \frac{1}{2} \eta^\top \tilde{C}_{i,j} \eta \\
        & - \eta^\top \big(x_j - \tilde{A}_{i, j} x_i - \tilde{b}_{i,j}\big),
    \end{aligned}
\end{equation}
where $i < j$. Following \cite{amatucci2025primal}, we can define the associative operator $\otimes$ for combining $V_{i\rightarrow k}$ and $V_{k\rightarrow j}$, when both are of the form in \eqref{eq:CVF}, to produce $V_{i\rightarrow j}$. The constituent terms of $V_{i\rightarrow j}$, as seen in \eqref{eq:CVF}, are obtained according to the following combination rules:
\begin{subequations}\label{eq:combination_rules}
\begin{align}
\tilde{P}_{i,j} &= \tilde{P}_{i,k} \otimes \tilde{P}_{k,j} \label{eq:comb_P}
\coloneqq \Upsilon_{i,j}\tilde{P}_{k,j} \tilde{A}_{i,k} + \tilde{P}_{i,k}, \\
\tilde{p}_{i,j} &= \tilde{p}_{i,k} \otimes \tilde{p}_{k,j} 
\coloneqq \Upsilon_{i,j} \big(\tilde{p}_{k,j} - \tilde{P}_{k,j} \tilde{b}_{i,k} \big) + \tilde{p}_{i,k}, \label{eq:comb_p}\\
\tilde{A}_{i,j} &= \tilde{A}_{i,k} \otimes \tilde{A}_{k,j} \label{eq:comb_A}
\coloneqq \Psi_{i,j} \tilde{A}_{i,k}, \\
\tilde{C}_{i,j} &= \tilde{C}_{i,k} \otimes \tilde{C}_{k,j} \label{eq:comb_C}
\coloneqq \Psi_{i,j} \tilde{C}_{i,k}\tilde{A}_{k,j}^\top + \tilde{C}_{k,j}, \\
\tilde{b}_{i,j} &= \tilde{b}_{i,k} \otimes \tilde{b}_{k,j} \label{eq:comb_b}
\coloneqq \Psi_{i,j} \big(\tilde{b}_{i,k} - \tilde{C}_{i,k}\tilde{p}_{k,j}\big) + \tilde{b}_{k,j}, \\
\text{where}&\quad
\Upsilon_{i,j} = \tilde{A}_{i,k}^\top \big(I + \tilde{P}_{k,j}\tilde{C}_{i,k}\big)^{-1}, \label{eq:comb_rule_upsilon} \\
\phantom{\text{where}}& \quad\Psi_{i,j} = \tilde{A}_{k,j}\big(I + \tilde{C}_{i,k}\tilde{P}_{k,j}\big)^{-1}. \label{eq:comb_rule_psi}
\end{align}
\end{subequations}
The initial values of the associative scan are defined as
\begin{subequations} \label{eq:assoc_init}
\hspace*{-0.4em}
\renewcommand{\arraystretch}{1.3}
    \begin{align}
        \tilde{P}_{i, i+1} = \hat{Q}_i - \hat{S}_i^\top \hat{R}_i^{-1}\hat{S}_i, &\quad \tilde{p}_{i, i+1} = \hat{q}_i - \hat{S}_i^\top \Omega_i \\
        \tilde{A}_{i, i+1} = A_i - B_i\hat{R}_i^{-1}\hat{S}_i, &\quad \tilde{C}_{i, i+1} = B_i\hat{R}_i^{-1}B_i^\top, \\
        \tilde{b}_{i, i+1} = b_i - B_i\Omega_i, &\quad \Omega_i = \hat{R}_i^{-1} \hat{r}_i \quad \forall i \in [N]\label{eq:comb_init} \\
        \tilde{P}_{N, N+1} = \hat{Q}_N, &\quad \tilde{p}_{N, N+1} = \hat{q}_N, \\
        \tilde{A}_{N, N+1} = 0,\quad \tilde{C}&_{N, N+1} = 0, \quad \tilde{b}_{N, N+1} = 0.
    \end{align}
\end{subequations}
As shown by \cite{amatucci2025primal}, using the CVF \eqref{eq:CVF} and combination rules \eqref{eq:combination_rules} with initialization \eqref{eq:assoc_init}, we can recover the Riccati terms $P_i = \tilde{P}_{i, N+1} \quad p_i = \tilde{p}_{i, N+1}$ through a reverse parallel associative scan \eqref{eq:revere_assoc_scan}. We can then recover the feedback terms

\vspace{-15pt}
\begin{subequations} \label{eq:controller_cache}
    \begin{align}
         K_i &= -\Gamma_i\big(\hat{S}_i + B_i^\top P_{i+1}A_i\big), \\
         k_i &= -\Gamma_i\left(B_i^\top(p_{i+1} + P_{i+1}b_i) + \hat{r}_i\right), \\
         \text{where}&\quad \Gamma_i = (\hat{R}_i + B_i^\top P_{i+1}B_i)^{-1}. \label{eq:gamma_cache}
    \end{align}
\end{subequations}
To recover the nominal updates $\delta x, \delta u$, we can use a \textit{forward} parallel associative scan to similarly parallelize the procedure. We define the conditional optimal trajectory (COT) as
\begin{equation} \label{eq:CVF_nominal}
    \bar{h}_{i \rightarrow j} (\delta x_i, \delta x_j) = \bar{A}_{i,j} \delta x_i + \bar{b}_{i,j}
\end{equation}
and using the combination rules with associated initialization,
\begin{equation}
\hspace*{-1.2em}
\renewcommand{\arraystretch}{1.3}
    \begin{array}{l@{\quad}l@{\quad}l}
        \bar{A}_{i,j} = \bar{A}_{i,k} \bar{A}_{k,j} & \bar{b}_{i,j} = \bar{A}_{k,j}\bar{b}_{i,k} + \bar{b}_{k,j},& \\
        \bar{A}_{i, i+1} = A_i + B_i K_i & \bar{b}_{i, i+1} = B_ik_i + b_i, & \forall i \in [1,N] \\
        \bar{A}_{0,1} = 0 & \bar{b}_{0,1} = A_0 \delta x_0 + b_0,
    \end{array}\hspace{-10pt}
\end{equation}
we can recover the solution to \eqref{eq:primal_update} in parallel by
\begin{equation} \label{eq:nom_recovery}
    \begin{aligned}
        &\delta x_i = \bar{h}_{0 \rightarrow 1} \otimes \bar{h}_{1 \rightarrow 2} \otimes \cdots \otimes \bar{h}_{i - 1 \rightarrow i} \\
        &\delta u_i = K_i \delta x_i + k_i.
    \end{aligned}
\end{equation}

Therefore, the solution to \eqref{eq:primal_update} can be calculated in $\mathcal{O} (\log N \log^2 n_x + \log^2 n_u)$ time on parallel hardware. This follows because each step of the reverse parallel associative scan requires inverting an $n_x \times n_x$ matrix, which can be performed in $\mathcal{O}(\log^2n_x)$ time in parallel \cite{csanky1975fast}. Since the scan has depth $\mathcal{O}(\log N)$, the total cost of the scan is $\mathcal{O}(\log N \log^2 n_x)$. The initialization step additionally requires matrix inversions of size $n_x$ and $n_u$, contributing $\mathcal{O}(\log^2 n_x + \log^2 n_u)$.

\subsubsection{Caching to Accelerate Associative Scans}
A common strategy to improve ADMM convergence is to update the penalty parameter $\rho$ only every $\sigma\in\mathbb{N}_{>1}$ iterations instead of every iteration. We adopt a similar $\rho$ update scheme as OSQP \cite{stellato2020osqp} presented in Section~5.2. By exploiting this structure, we cache and reuse matrix factorizations across iterations in which $\rho$ remains fixed, substantially reducing computational cost.

\looseness-1To solve \eqref{eq:primal_update} when $\rho$ is fixed, only the augmented linear terms in \eqref{eq:augmented_linear} change between ADMM iterations, i.e., $\hat Q_{(\cdot)}$, $\hat R_{(\cdot)}$, and $\hat S_{(\cdot)}$ are unchanged. Thus, in the reverse parallel associative scan \eqref{eq:combination_rules}, only the quantities $\tilde{p}_{i,j}$ \eqref{eq:comb_p} and $\tilde{b}_{i,j}$ \eqref{eq:comb_b} must be recomputed. Since the dominant computational cost arises from matrix inversions, we also cache the invariant intermediate quantities $\tilde{P}_{i,j}$ \eqref{eq:comb_P}, $\tilde{C}_{i,j}$ \eqref{eq:comb_C}, $\Upsilon_{i,j}$, and $\Psi_{i,j}$ \eqref{eq:comb_rule_upsilon}-\eqref{eq:comb_rule_psi}, which are needed to compute $\tilde{p}_{i,j}$ and $\tilde{b}_{i,j}$, so that they can be reused across iterations.
Therefore, for ADMM iterations in which $\rho$ is unchanged, the combination rules \eqref{eq:combination_rules} reduce to
\begin{equation} \label{eq:comb_cache}
    \begin{aligned}
        \tilde{p}_{i,j} &= \Upsilon_{i,j}^{\text{cache}} \big(\tilde{p}_{k,j} - \tilde{P}_{k,j}^{\text{cache}} \tilde{b}_{i,j}\big), \\
        \tilde{b}_{i,j} &= \Psi_{i,j}^{\text{cache}} \big(\tilde{b}_{i,k} - \tilde{C}_{i,k}^{\text{cache}} \tilde{p}_{i,j} \big) + \tilde{b}_{j,k},
    \end{aligned}
\end{equation}
where $ \Upsilon_{i,j}^{\text{cache}}, \tilde{P}_{k,j}^{\text{cache}}, \Psi_{i,j}^{\text{cache}}, \tilde{C}_{i,k}^{\text{cache}}$ are the cached terms from the $\rho$-updated iteration \eqref{eq:combination_rules}. The initial values are set as
\begin{equation}
\hspace*{-0.0em}
\renewcommand{\arraystretch}{1.5}
    \begin{array}{l@{\quad}l@{\quad}l}
        \tilde{p}_{i, i+1} = \hat{q}_i - \hat{S}_i \Omega_i^{\text{cache}}, & \tilde{b}_{i,i+1} = b_i - B_i \Omega_i^{\text{cache}}, & \forall i \in [N] \\
        \tilde{p}_{N, N + 1} = \hat{q}_N, & \tilde{b}_{N, N+1} = 0,
    \end{array}
\end{equation}
where $\Omega_i^{\text{cache}}$ are the cached term from \eqref{eq:comb_init}. We can then obtain the feedback terms $K_i$ and $k_i$ as
\begin{equation} \label{eq:gains_cache}
    \begin{aligned}
        \hspace{-6pt}& K_i = K_i^\text{cache}\hspace{-3pt}, 
        \ \ k_i = -\Gamma_i^{\text{cache}} \left( B_i^\top (p_{i+1} + P_{i+1}^{\text{cache}}b_i) + \hat{r}_i\right)\hspace{-7pt}
    \end{aligned}
\end{equation}
where $\Gamma_i^{\text{cache}}$ is the cached term from \eqref{eq:gamma_cache}. Using these feedback terms, we can follow the same procedure \eqref{eq:CVF_nominal}--\eqref{eq:nom_recovery} to recover the nominal update $\delta x, \delta u$. Because procedures \eqref{eq:comb_cache}--\eqref{eq:gains_cache} and \eqref{eq:CVF_nominal}--\eqref{eq:nom_recovery} require no inverses, we perform one iteration in $\mathcal{O}(\log N \log n_x + \log n_u)$, a reduction from $\mathcal{O} (\log N \log^2 n_x + \log^2 n_u)$ without caching. 

\subsection{Robust Nonlinear MPC via System Level Synthesis}\label{sec:method_sls}
To solve for the robust SLS controller \eqref{eq:fastsls_qp}, FastSLS uses $N$ independent Riccati recursions, as shown in \cite{leeman2024fast}:
\begin{equation} \label{eq:backward_prop}
    \begin{aligned}
        \mathcal{G}_{N}^{(j)} &= \mathcal{Q}^{\text{x}}_{N,j}, \quad \quad 
        \mathcal{K}_{k}^{(j)} = -\mathcal{G}_{k}^{(j)} \mathcal{B}_{k}^{(j)}, \\
        \mathcal{P}_{k}^{(j)} &= \mathcal{Q}^\text{x}_{k,j} + \big(A^{(s)}_k\big)^\top \mathcal{P}_{k+1}^{(j)}A^{(s)}_k + \big(\mathcal{K}_{k}^{(j)}\big)^\top\mathcal{B}_{k}^{(j)}, \\
        \mathcal{G}_{k}^{(j)} &= \big(\mathcal{Q}^{\text{u}}_{k,j} + \big(B^{(s)}_k\big)^\top\mathcal{P}_{k+1}^{(j)}B^{(s)}_k\big)^{-1}, \\
        \mathcal{B}_{k}^{(j)} &= \mathcal{Q}^{\text{ux}}_{k,j} + \big(B_k^{(s)}\big)^\top \mathcal{P}_{k+1}^{(j)} A^{(s)}_k,
    \end{aligned}
\end{equation}
and $N$ independent forward propagations
\begin{equation} \label{eq:forward_prop}
    \hspace{-6pt}\begin{aligned}
         &\mathbf{\Phi}^{\text{x}}_{j+1,j} = E_j, \\
         &\mathbf{\Phi}^u_{k,j} = \mathcal{K}_{k}^{(j)} \mathbf{\Phi}^{\text{x}}_{k,j}, & \mathbf{\Phi}^{\text{x}}_{k+1,j} = (A^{(s)}_k + B^{(s)}_k \mathcal{K}_{k}^{(j)})\mathbf{\Phi}^{\text{x}}_{k,j}, \\
    \end{aligned}\hspace{-7pt}
\end{equation}
for all $j \in [N]$ and for all $k \in [j+1,N-1]$, where $j$ indexes the time at which the disturbance is injected, and 
$k$ indexes the state and input responses at time $k$ for the corresponding disturbance. The superscript $(j)$ denotes the Riccati variables associated with the disturbance injected at time $j$.

We use parallel associative scans to calculate both \eqref{eq:backward_prop} and \eqref{eq:forward_prop} to get the solution for \eqref{eq:fastsls_qp}. For a given Riccati recursion for the $j$-th disturbance, we define the CVF
\begin{equation} \label{eq:CVF_robust}
    \begin{aligned}
        \mathcal{V}^{(j)}_{i \rightarrow l}(\mathbf{\Phi}_{i,j}, \mathbf{\Phi}_{l,j}) = \max_\theta &\frac{1}{2} \mathbf{\Phi}_{i,j}^\top\tilde{\mathcal{P}}_{i,l}^{(j)} \mathbf{\Phi}_{i,j} - \frac{1}{2} \theta^\top \tilde{\mathcal{D}}_{i,l}^{(j)} \theta \\
        & - \theta^\top \big(\mathbf{\Phi}_{l,j} - \tilde{\mathcal{A}}_{i, l}^{(j)} \mathbf{\Phi}_{i, j}\big),
    \end{aligned}
\end{equation}
and the associated combination rules as
\begin{equation}\label{eq:robust_combin}
    \begin{aligned}
        \tilde{\mathcal{P}}^{(j)}_{i,l} &= \big(\tilde{\mathcal{A}}^{(j)}_{i,k}\big)^\top\big(I + \tilde{\mathcal{P}}^{(j)}_{k,l} \tilde{\mathcal{D}}^{(j)}_{i,k}\big)^{-1}\tilde{\mathcal{P}}^{(j)}_{k,l}\tilde{\mathcal{A}}^{(j)}_{i,k} + \tilde{\mathcal{P}}^{(j)}_{i,k}, \\
        \tilde{\mathcal{A}}^{(j)}_{i,l} &= \tilde{\mathcal{A}}^{(j)}_{k,l}\big(I + \tilde{\mathcal{D}}^{(j)}_{i,k} \tilde{P}^{(j)}_{k,l}\big)^{-1} \tilde{A}^{(j)}_{i,k}, \\
        \tilde{\mathcal{D}}^{(j)}_{i,l} &= \tilde{\mathcal{A}}^{(j)}_{k,l}\big(I + \tilde{\mathcal{D}}^{(j)}_{i,k} \tilde{P}^{(j)}_{k,l}\big)^{-1} \tilde{\mathcal{D}}^{(j)}_{i,k}\big(\tilde{\mathcal{A}}^{(j)}_{k,l}\big)^\top + \tilde{\mathcal{D}}^{(j)}_{k,l}.
    \end{aligned}
\end{equation}
We define the initial values as
\begin{equation}
    \begin{aligned}
        \tilde{\mathcal{A}}_{i, i+1}^{(j)} &= A^{(s)}_i - B^{(s)}_i \big(\mathcal{Q}_{i,j}^{\text{u}}\big)^{-1}\mathcal{Q}^{\text{xu}}_{i,j}, \\
        \tilde{\mathcal{D}}_{i, i+1}^{(j)} &= B^{(s)}_i \big(\mathcal{Q}_{i,j}^{\text{u}}\big)^{-1} \big(B^{(s)}_i\big)^\top, \\
    \tilde{\mathcal{P}}_{i,i+1}^{(j)} &= \mathcal{Q}^{\text{x}}_{i,j} - \mathcal{Q}^{\text{ux}}_{i,j} \big(\mathcal{Q}^{\text{u}}_{i,j}\big)^{-1} \mathcal{Q}^{\text{xu}}_{i,j}, \\
    \tilde{\mathcal{P}}^{(j)}_{N, N + 1} = &\mathcal{Q}^{\text{x}}_{N+1}  \quad  \tilde{\mathcal{A}}^{(j)}_{N, N+1} = 0, \quad \tilde{\mathcal{D}}^{(j)}_{N, N + 1} = 0.
    \end{aligned}
\end{equation}
We can perform a reverse parallel associative scan \eqref{eq:revere_assoc_scan} using combination rules and initialization \eqref{eq:robust_combin} for each disturbance $E_j$ independently in parallel to recover the Riccati terms by $\mathcal{P}_k^{(j)} = \tilde{\mathcal{P}}^{(j)}_{k, N + 1}$. Using the backward pass terms \eqref{eq:backward_prop}, we can recover all Riccati feedback gains $\mathcal{K}_{k}^{(j)}$. To retrieve the forward pass terms \eqref{eq:forward_prop}, we follow a similar procedure as the forward propagation of the nominal trajectory \eqref{eq:CVF_nominal}--\eqref{eq:nom_recovery}. We define the Conditional Optimal Propagation (COP) $\bar{\mathcal{N}}_{i \rightarrow l} (\mathbf{\Phi}^x_{i, j}, \mathbf{\Phi}^x_{l, j}) = \bar{\mathcal{A}}_{i,l}^{(j)} \mathbf{\Phi}^{\text{x}}_{i,j}$ with combination rule and initialization
\begin{equation}
    \begin{aligned}
        \bar{\mathcal{A}}_{i, l} ^{(j)} = \bar{\mathcal{A}}_{i, k} ^{(j)}\bar{\mathcal{A}}_{k, l} ^{(j)}, \qquad \bar{\mathcal{A}}_{i,i+1}^{(j)} = A_i^{(s)} + B_i^{(s)}\mathcal{K}_{i,j},
    \end{aligned}
\end{equation}
for $i = 1,\ldots, N$, while for $i = 0$ we set $\bar{\mathcal{A}}_{0,1} = 0$. From the associative scan, we obtain the solution to \eqref{eq:fastsls_qp} by
 \begin{equation}
    \begin{aligned}
        \mathbf{\Phi}^{\text{x}}_{i,j} &= \bar{\mathcal{N}}_{0 \rightarrow 1} \otimes \ldots \otimes \bar{\mathcal{N}}_{i-1 \rightarrow i}, \qquad \mathbf{\Phi}^{\text{u}}_{i,j} = \mathcal{K}_{i,j} \mathbf{\Phi}^{\text{x}}_{i,j}.
    \end{aligned}
\end{equation}
Therefore, we similarly can calculate the robust control policy in $\mathcal{O} (\log N \log^2 n_x + \log^2 n_u)$ time on parallel hardware.

\subsection{Real-Time Iteration} \label{sec:rti}
As standard practice for achieving real-time performance in MPC, we employ a Real-Time Iteration (RTI) scheme in Fig.~\ref{fig:combined_quad}. In RTI, only a single SQP iteration is performed per control step, sacrificing linearization accuracy for computational speed. This approximation is typically acceptable, as subsequent MPC updates relinearize the dynamics around the newly executed state, thereby correcting the linearization error. By only performing one iteration, this drastically improves the computation time for one control, making these MPC loops suitable for real-time.

We adopt an RTI framework for our solver, resulting in the RTI GPU-SLS solver. In this setting, we modify our algorithm to instead perform a single linearization of the dynamics and constraints. We warm-start the controller update using the previous iteration's $\tau$ variables, from which we can calculate the new constraint tightenings. These are then used to solve the tightened nominal trajectory optimization, which is applied for the current control step. Therefore, the total solve time for one RTI update consists of the required time for one linearization, one controller update, and one tightened nominal trajectory update.

\section{Results}\label{sec:results}
\begin{figure}[t] \label{figure:solve_comparisons}
  \centering
  \includegraphics[width=\linewidth, trim=0 0 0 40,
  clip]{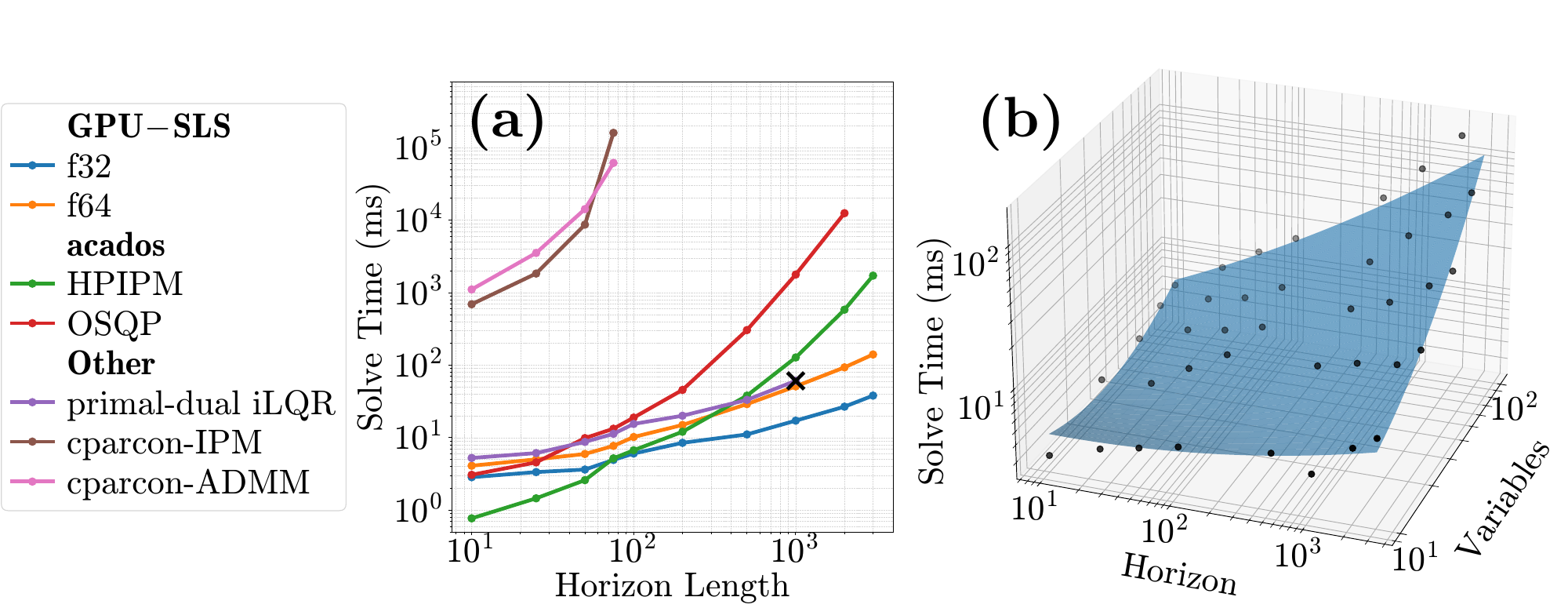} \vspace{-15pt}
  \caption{\textbf{(a)}: Comparison of average solve-time scaling with horizon length for various NMPC solvers on a torque-constrained 10-link pendulum stabilization task under random external disturbances. Our method consistently outperforms all baselines for long-horizon problems. \textbf{(b)}: Solve-time scaling of GPU-SLS with a family of $n$-link torque-constrained pendulums.}
  \label{fig:solver-comparison} \vspace{-15pt}
\end{figure}
In this section, we evaluate the computational efficiency of our nonlinear MPC framework against state-of-the-art CPU and GPU-based solvers (Sec.~\ref{sec:nmpc_benchmarks}), assess the robustness and speedups of our proposed SLS solver against existing methods (Sec.~\ref{sec:sls_benchmarks}), and demonstrate the practical applicability of our work through both simulation and hardware experiments on legged locomotion systems (Sec. \ref{sec:legged_expe}). All CPU benchmarks were run on a desktop computer equipped with an \textit{AMD Ryzen 9900X} processor (12 cores, 24 threads). GPU-based hardware experiments were conducted on a system with an \textit{NVIDIA RTX 4070 Ti Super}, while all other remaining GPU simulation experiments were conducted using a \textit{NVIDIA RTX 4090}.
\subsection{Constrained Nonlinear MPC Benchmarks} \label{sec:nmpc_benchmarks}
To demonstrate the scalability of our formulation to long horizons and large problem sizes, we compare our method against various constrained NMPC solvers on a torque-constrained 10-link inverted pendulum. At each control step, the system is randomly perturbed at each joint, requiring the NMPC to stabilize the system. All benchmarks were solved to convergence to a maximum residual tolerance of $10^{-2}$. Reported runtimes correspond to the average solve time over 1000 MPC iterations.

As shown in Fig.~\ref{fig:solver-comparison}, the proposed method substantially outperforms CPU-based solvers OSQP and HPIPM, reducing solve times by $\textbf{99.8\%}$ and $\textbf{98.9\%}$, respectively, in long-horizon scenarios. Our method also improves upon the GPU-accelerated augmented Lagrangian approach (primal-dual iLQR AL) by $\textbf{71.8\%}$ and  significantly outperforms both \textit{cparcon-IPM} and \textit{cparcon-ADMM}. Furthermore, we evaluate the impact of using high precision arithmetic (FP64) to handle occasionally ill-conditioned problems and observe that our approach continues to outperform all baselines.

These scaling trends can be attributed to the computational structure of each solver. CPU-based methods such as HPIPM and OSQP rely on largely sequential linear-algebra operations, which becomes a bottleneck as the horizon grows. Primal-dual iLQR AL, while GPU-parallelized, incurs higher per-iteration cost due to increased sensitivity of the penalty parameter under reduced GPU precision, and in certain ill-conditioned problems, will lead to divergence, as reflected in our experiments. To improve convergence, \textit{cparcon} uses full second-order information rather than a Gauss-Newton approximation, leading to multiple expensive associative scan operations that increases wall-clock time. ADMM on the other hand, can tolerate inexact solves on the GPU more reliably than AL methods. This attribute, paired with our caching parallel associative scan routine, enables substantial performance improvements.

We also show in Fig.~\ref{fig:solver-comparison} that we are able to solve a family of torque-constrained $n$-link pendulum problems with horizons as large as $N = 3000$, corresponding to more than $1.35 \times 10^5$ decision variables within 73 milliseconds. These results reflect our formulation's logarithmic scaling with respect to the state, control, constraint, and horizon dimensions, allowing our method to solve substantially larger problems beyond existing MPC methods.

\subsection{SLS Benchmarks} \label{sec:sls_benchmarks} 
\begin{figure}[t]
  \centering
  \includegraphics[width=\linewidth]{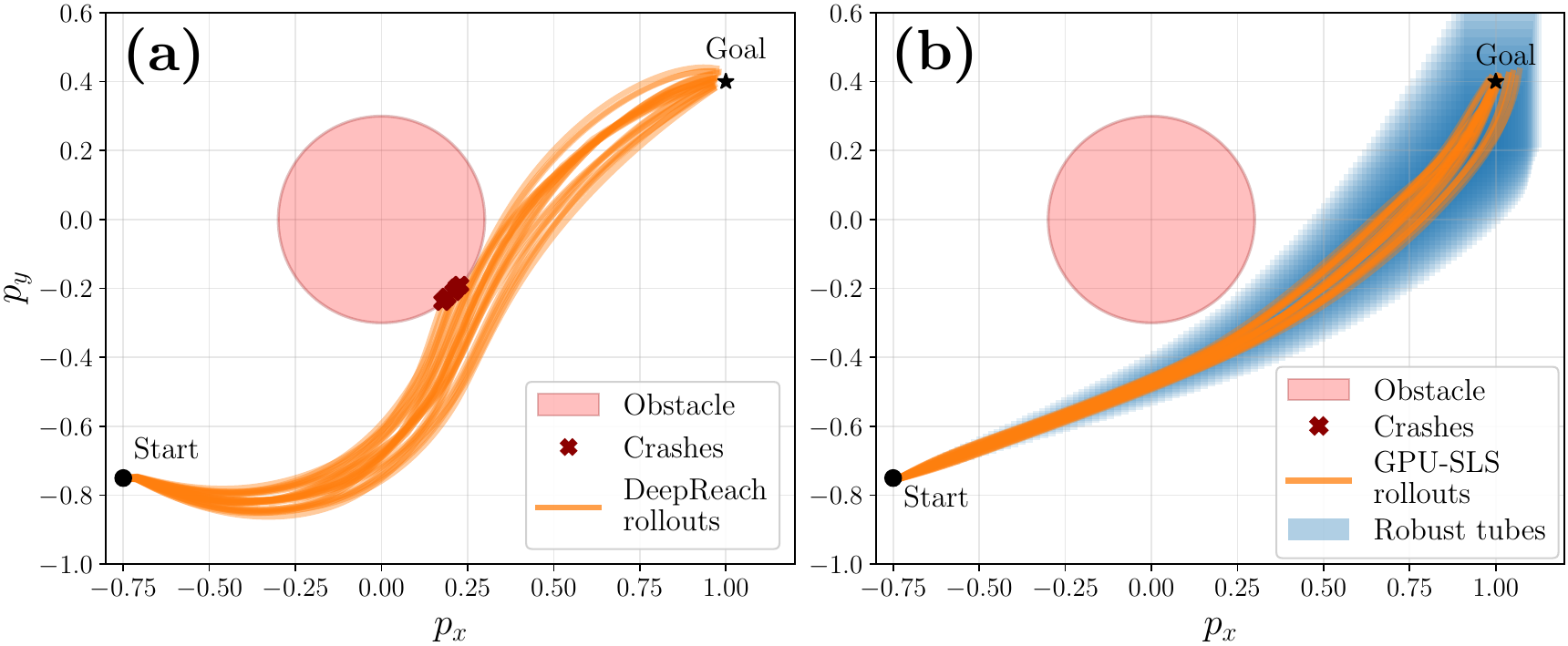} \vspace{-20pt}
  \caption{DeepReach rollouts \textbf{(a)} and GPU-SLS rollouts \textbf{(b)} for an obstacle avoidance Dubins car system subject to both adversarial and random disturbance. Deepreach is shown to fail to consistently certify safety, while our method certifies safety 100\% of all rollouts.}
  \label{fig:deepreach}  \vspace{-5pt}
\end{figure}

\begin{figure}[t]
  \centering
  \includegraphics[width=\linewidth,trim=180pt 6pt 180pt 8pt,
  clip]{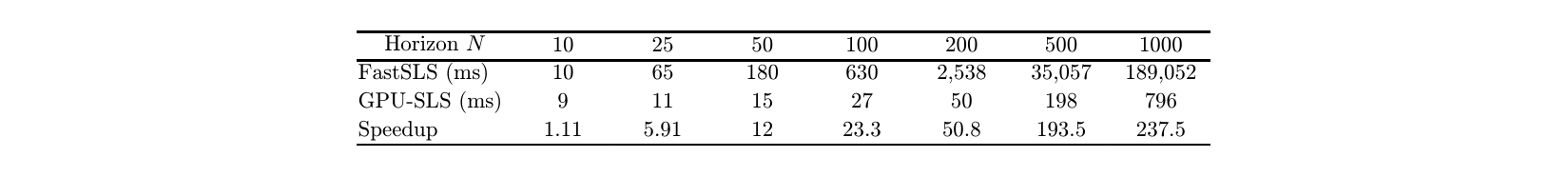} \vspace{-20pt}
  \caption{Runtime comparison of FastSLS vs GPU-SLS for solving the system in \ref{fig:deepreach}. GPU-SLS consistently outperforms FastSLS due to its logarithmic scaling vs FastSLS's cubic scaling.}
  \label{fig:runtim_table}  \vspace{-15pt}
\end{figure}

\begin{figure}[t]
  \centering
  \includegraphics[width=\linewidth]{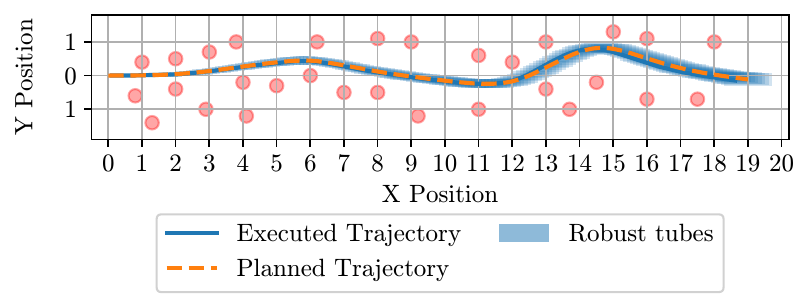} \vspace{-23pt}
  \caption{Dubins car experiment using GPU-SLS to calculate a robust trajectory for 20 meters through a dense constraint field of 30 obstacles. This demonstrates our solver's ability to handle large-scale trajectory optimization and robust constraint satisfaction at scale.}
  \label{fig:big_dubins} \vspace{-5pt}
\end{figure}
\begin{figure}[t] \vspace{-5pt}
  \centering
    \includegraphics[width=\linewidth, trim={0 210pt 0 0}, clip]{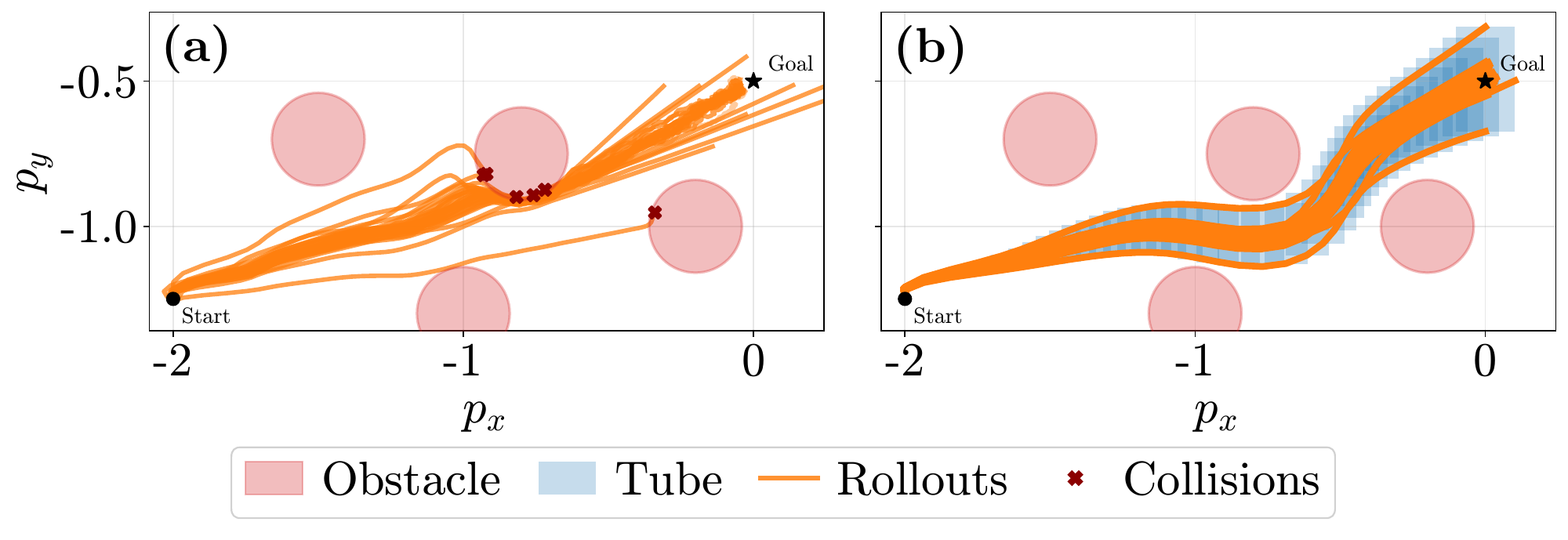} \vspace{-23pt}
  \caption{Rollouts for uncertain nonlinear dynamics on a 6D planar quadrotor using \textbf{(a)}: Classical Closed-Loop MPC and \textbf{(b)}: robust MPC via GPUSLS. Classical MPC fails to reliably avoid the obstacle, whereas our method consistently achieves safe navigation.}
  \label{fig:classical_mpc_comparison} \vspace{-20pt}
\end{figure}
In this section, we evaluate our GPU-SLS on a suite of Dubins car benchmarks to assess safety, scalability, and computational performance. We further compare the benefits of robust MPC against classical MPC on a planar quadrotor. The associated state, control, and dynamics definitions are provided in App.~\ref{app:dubins car}--\ref{app:planar_quadrotor}.

To demonstrate the superior safety-rating of our method compared to data-driven methods such as DeepReach, we perform closed-loop rollouts \eqref{eq:disturbance_feedback} of a Dubins car system navigating around an obstacle. In Fig.~\ref{fig:deepreach}, we plot the rollouts of both controllers subject to random and adversarial disturbance. DeepReach fails to consistently certify safety, crashing into the obstacle in 6 out of 31 rollouts. This is because DeepReach relies on offline value-function approximations, which requires large-amounts of training data and provides limited robustness guarantees. GPU-SLS, on the other hand, explicitly enforces robustness through disturbance-feedback synthesis and certifies safety in all tested rollouts, highlighting the limitations of purely data-driven safety certificates.

We further demonstrate the advantages of robust MPC over classical MPC on a 6D planar quadrotor. Under disturbances, standard MPC (Fig. \ref{fig:classical_mpc_comparison}\textbf{a}) fails to consistently avoid all obstacles, colliding in 6/31 executions, whereas our method (Fig. \ref{fig:classical_mpc_comparison}\textbf{b}) successfully avoids all obstacles across all rollouts. Notably, the robustness procedure adds minimal overhead, with the controller solve accounting for $\sim$3\% of total computation time. 

A core bottleneck with traditional FastSLS is its computational speed. In Fig.~\ref{fig:runtim_table}, we benchmark the average solve times of GPU-SLS and FastSLS on a similar system as in Fig.~\ref{fig:deepreach}. As the prediction horizon increases, our method achieves progressively larger speedups compared to FastSLS, with the largest horizon offering a $237.5 \times$ speedup. This contrast reflects FastSLS' cubic scaling with respect to problem dimensions, whereas our parallelized routine exhibits a logarithmic scaling across the same dimensions, allowing our formulation to solve robust OCPs at scale.

We further evaluate the scalability of our formulation to large environments with multiple constraints. Fig.~\ref{fig:big_dubins} illustrates a long horizon Dubins car trajectory (20 m) navigating through a dense constraint field of 30 obstacles subject to external disturbances. The system is able to retain safety while navigating difficult terrain, consistently remaining within the calculated reachable tubes. These results demonstrate the ability of GPU-SLS to handle large-scale, highly constrained problems without compromising robustness.

\subsection{Legged Experiments} \label{sec:legged_expe}
\begin{figure}[t]
  \centering
  \includegraphics[width=\linewidth, trim=0 200 350 0, clip]{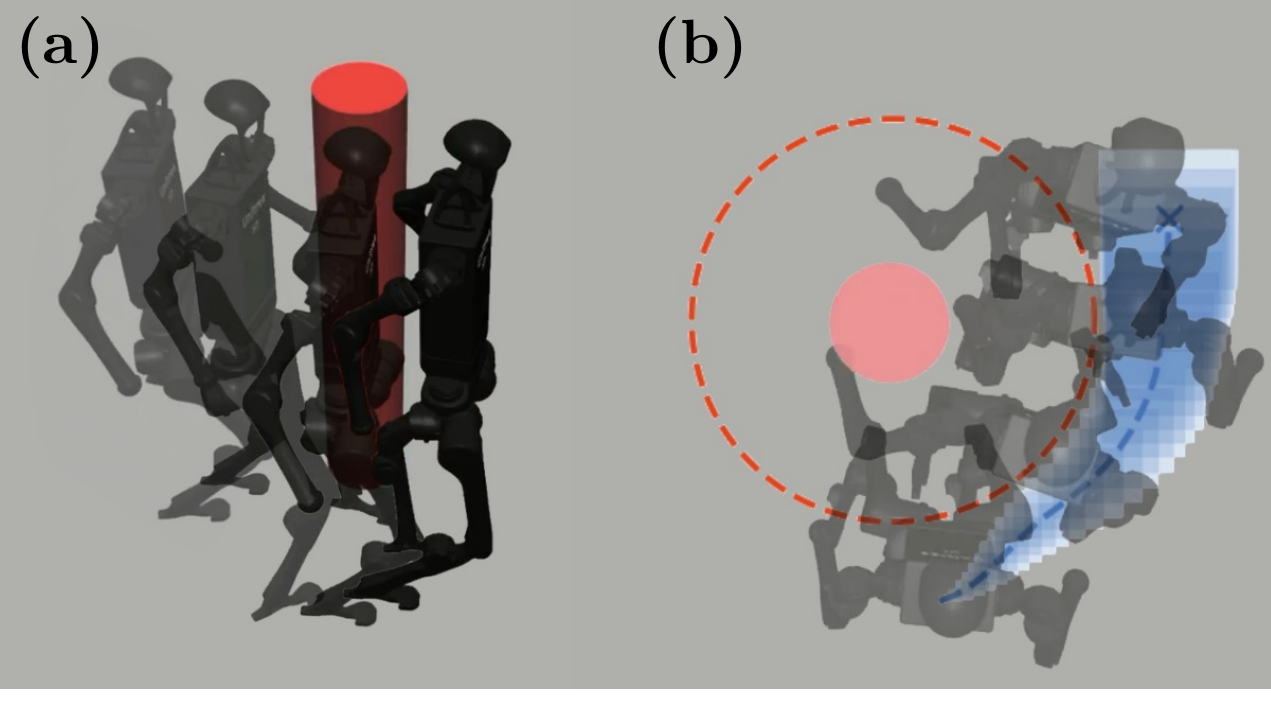} \vspace{-20pt}
  \caption{\textbf{(a)}: Whole-body humanoid (75D, 19C) navigating around an obstacle using \eqref{eq:disturbance_feedback} calculated from GPU-SLS. \textbf{(b)}: A visualization of the robust forward tubes (blue squares) generated by \eqref{eq:fastsls_problem} and the humanoid's rolled out trajectory remaining within the tubes.}
  \label{fig:humanoid_timelapse} \vspace{-5pt}
\end{figure}

\begin{figure}[t]
  \centering
  \includegraphics[width=\linewidth]{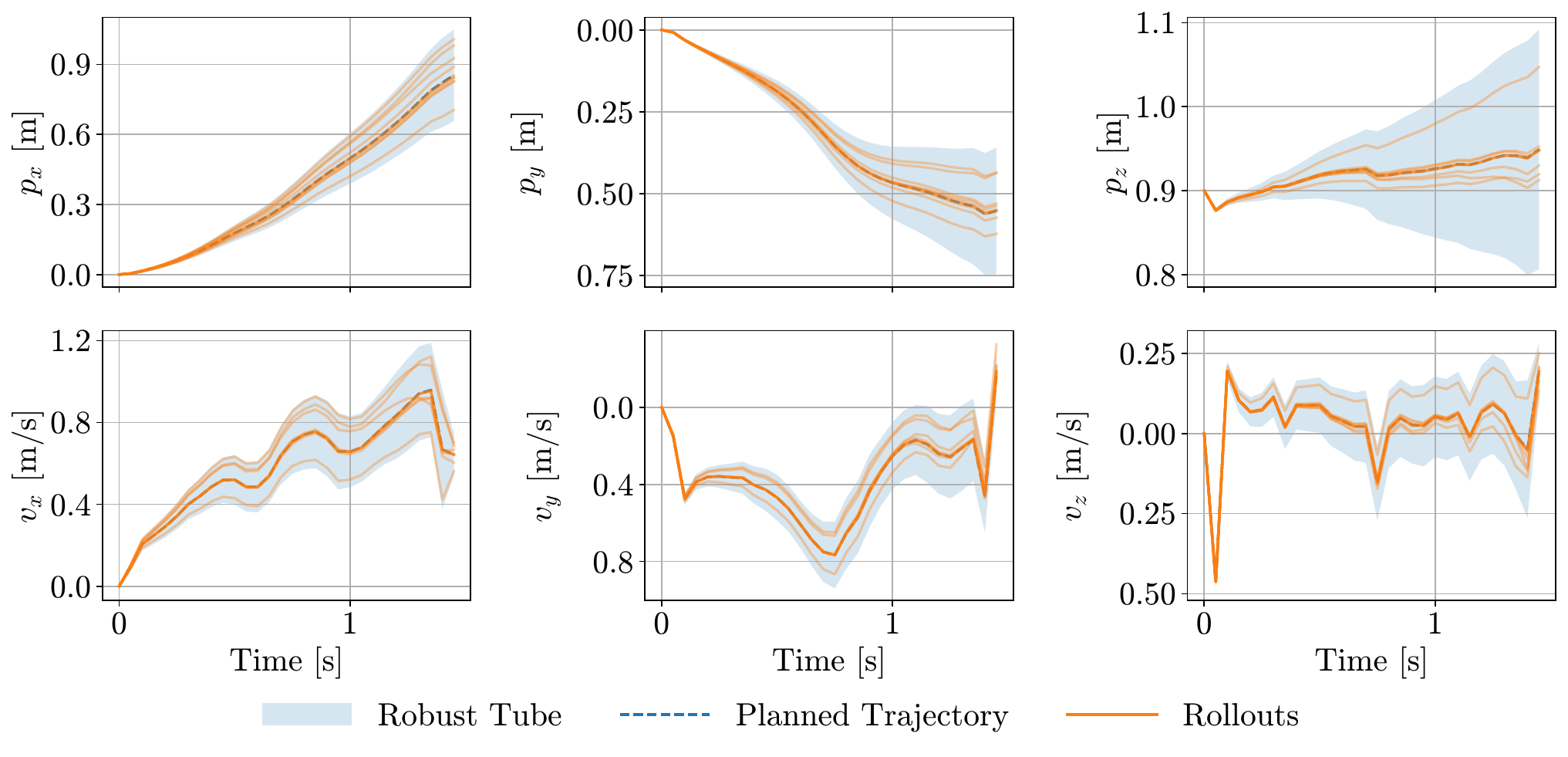} \vspace{-20pt}
  \caption{Simulated rollouts of \ref{fig:humanoid_timelapse} under adversarial disturbances. Using the controller \eqref{eq:disturbance_feedback} the humanoid is able to safely remain within the tubes, demonstrating our method's ability to develop robust control policies for complex, high-dimensional nonlinear systems.}
  \label{fig:humanoid_tubes} \vspace{-5pt}
\end{figure}

\begin{figure}[t]
  \centering
  \includegraphics[
  width=\linewidth,
  trim=0pt 8pt 0pt 22pt,
  clip
]{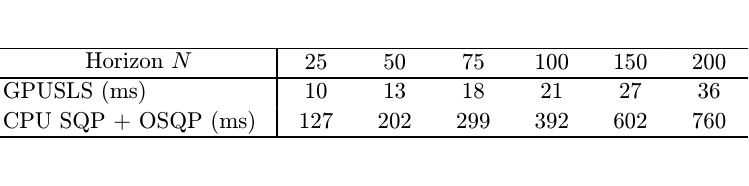} \vspace{-29pt}

  \caption{Runtime comparisons between our GPU solver and the CPU-based solver OSQP for varying horizons on a whole rigid-body quadruped (61D) navigating around an obstacle. Across all horizons, our method outperforms OSQP, achieving up to a 95\% reduction in solve time at the longest horizon.}
  \label{fig:solve_comparisons} \vspace{-2pt}
\end{figure}

\begin{figure}[t]
  \centering
  \includegraphics[width=\linewidth, trim=0 250 600 0, clip]{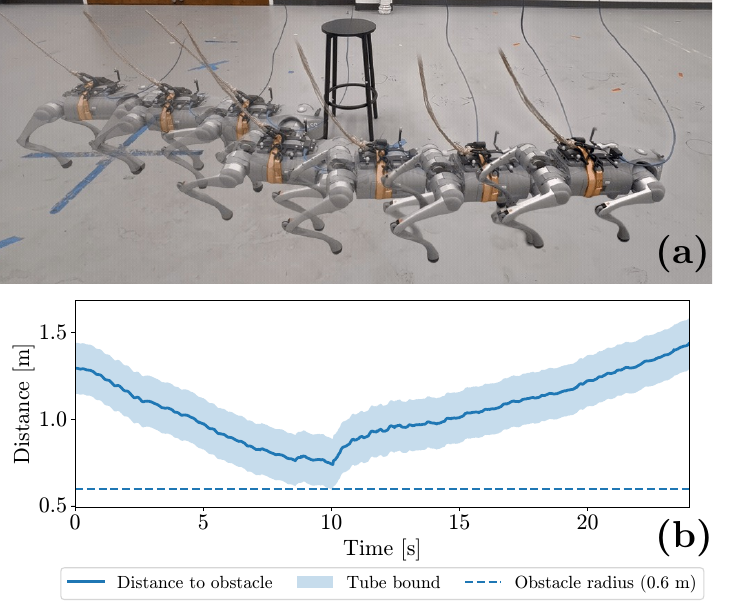}
  \caption{\textbf{(a)}: GPU-accelerated robust whole-body control (61 states, 12 controls) using RTI GPU-SLS executed on hardware with a Unitree Go2 EDU quadruped. The robot successfully navigates robustly around an obstacle in real-time at 50 Hz. \textbf{(b)}: Graph illustrating the robot's distance to the obstacle and the maximum tube size for each time step along the trajectory. The tubes are used to tighten the constraints in the nominal trajectory generator, ensuring that the nominal policy remains safe despite disturbances.}\vspace{-10pt}
  \label{fig:combined_sls_quad}
\end{figure}
We evaluate our solver's ability to plan robust whole-body trajectories for a Unitree H1 humanoid robot with 75 states and 19 control inputs in a MuJoCo simulation environment adapted from \cite{amatucci2025primal} using the humanoid's full, rigid-body dynamics model. As demonstrated in Fig.~\ref{fig:humanoid_timelapse}, we task the humanoid with robustly navigating around an obstacle subject to external disturbances. The humanoid can be seen successfully navigating around the obstacle using the synthesized robust controller \eqref{eq:disturbance_feedback}, which ensures that the system state remains within its robust tubes as shown in Fig.~\ref{fig:humanoid_tubes}. Due to the parallelization of the GPU and the formal guarantees of SLS, we are able to solve constrained, high dimensional nonlinear systems, while also calculating robust controllers that guarantee the safety of the system.

The high computation times of traditional constrained NMPC solvers preclude their applicability to legged hardware experiments requiring high control update rates. To address this, we first benchmark the computation times between CPU-based solver OSQP and our solver for controlling a whole rigid-body quadruped (61D) in navigating around an obstacle. Both methods are implemented in an RTI scheme and solved to the same convergence tolerance of $10^{-2}$, with solve times averaged across 700 MPC iterations. In Fig.~\ref{fig:solve_comparisons} we show the benefit of our GPU accelerated method, outperforming OSQP in all horizon lengths, achieving a 95\% speedup at the longest horizon, while requiring fewer ADMM iterations (27/25 mean/median vs. 107/50 for OSQP). The difference in computation time stems from the parallel structure of our method, which is particularly advantageous for high-dimensional systems and longer horizons, enabling real-time performance where CPU-based solvers become computationally prohibitive.

To validate the real-world practicality of our formulation for legged robots, we perform whole-body constrained NMPC and SLS on a physical Unitree Go2 EDU quadruped (61D). In our hardware experiments, we use a Vicon Motion Capture System to track the quadruped's center of mass. In the first experiment, shown in Fig.~\ref{fig:combined_quad}, the quadruped is tasked with safely navigating through two obstacles using the nominal trajectory generator. The controller runs at 50 Hz with an average solve time of 16ms, prediction horizon of 25, and a discretization timestep of 0.02s. From Fig.~\ref{fig:combined_quad} we can see that the quadruped successfully navigates through the obstacles in a collision-free manner, highlighting the practical viability of the proposed constrained optimization framework on hardware. 

We also evaluate the practicality of the RTI GPU-SLS pipeline for real-time reachability analysis on hardware. We task the robot with navigating around an obstacle with a disturbance magnitude of 0.02. In Fig.~\ref{fig:combined_sls_quad}, we show that the quadruped successfully navigates around the obstacle while remaining within its tubes throughout the entire trajectory, demonstrating the practicality on hardware of our RTI GPU-SLS algorithm for high-dimensional nonlinear systems.

\section{Conclusion} 
\label{sec:conclusion}

We presented GPU-SLS, a GPU-parallelized framework for safe, large-scale robust NMPC that unifies constrained trajectory optimization and disturbance-feedback controller synthesis within a real-time pipeline. By exploiting the structure of ADMM and FastSLS, our method leverages parallel associative scans and factorization caching to achieve logarithmic-depth scaling in horizon, control, and state dimensions, enabling millisecond-scale solutions to robustly constrained optimization problems. Our method demonstrates significant speedups over existing CPU and GPU solvers, along with 100\% empirical safety across high-dimensional systems, including both simulation and hardware validation on humanoids and quadrupeds. Overall, GPU-SLS removes a computational bottleneck, enabling real-time, safe robust control for high-dimensional robotic systems.

\section*{Acknowledgments}

We thank Saman Zonouz and the Indoor Flight Laboratory at the Georgia Institute of Technology for support with hardware and experimental facilities.

\bibliographystyle{plainnat}
\bibliography{references}
\clearpage
\appendix
In this section, we provide supplemental material to support the information provided in our methodology. App.~\ref{app:parallel_associative_scan} discusses the mechanism of parallel associative scans, which are the critical component to our GPU acceleration. App.~\ref{app:fastsls_duals} discusses the meaning of the FastSLS dual variables $\tau$ and how they are calculated. App.~\ref{app:dubins car} and App.~\ref{app:planar_quadrotor} defines the canonical state space, control space, and dynamics of a Dubins car system and planar quadrotor respectively. App.~\ref{app:quad_and_human} lists the simulation environment for the quadruped and humanoid experiments, along with their associated state and control space. App.~\ref{app:humanoid_rti_ext} extends RTI GPU-SLS to a humanoid robot. Lastly, App.~\ref{app:humanoid_tubes_all_of_thm} displays the remaining state tubes and rollouts from Fig. \ref{fig:humanoid_timelapse}.
\appendices
\section{Parallel Associative Scan}\label{app:parallel_associative_scan}
This appendix briefly reviews the parallel associative scan algorithm and explains why it enables efficient parallelization in our formulation. We proceed with a generic explanation of the parallel associative scan algorithm. 

Consider a set of elements $\mathcal{S} = \{a_1, a_2, ..., a_n\}$ and an associative operator $\otimes$ such that $(a \otimes b) \otimes c = a \otimes (b \otimes c) \quad \forall a,b,c \in \mathcal{S}$. We want to find the following elements in $\mathcal{O}(\log n)$ complexity.
\begin{equation}
    \begin{aligned}
        s_1 &= a_1, \\
        s_2 &= a_1 \otimes a_2, \\
        s_3 &= a_1 \otimes a_2 \otimes a_3, \\
        \vdots & \\
        s_n &= a_1 \otimes a_2 \otimes \ldots \otimes a_n.
    \end{aligned} \label{eq:prefix_sums}
\end{equation}

Parallel associative scan is typically implemented using a balanced binary tree, where each leaf node corresponds, in left-to-right order, to the sequence of elements $\{a_1, a_2, \ldots, a_n\} \in \mathcal{S}$. The first phase of the algorithm is to perform an upsweep.

During the upsweep phase, the algorithm proceeds bottom-up through the tree, recursively combining pairs of child nodes using the associative operator $\otimes$ and storing the result at their parent node. We use a simple example of $n = 4$ to illustrate the process. 
\begin{figure}[h]
    \centering
    \includegraphics[width=\columnwidth,
    trim=0 870pt 0 0,
        clip
    ]{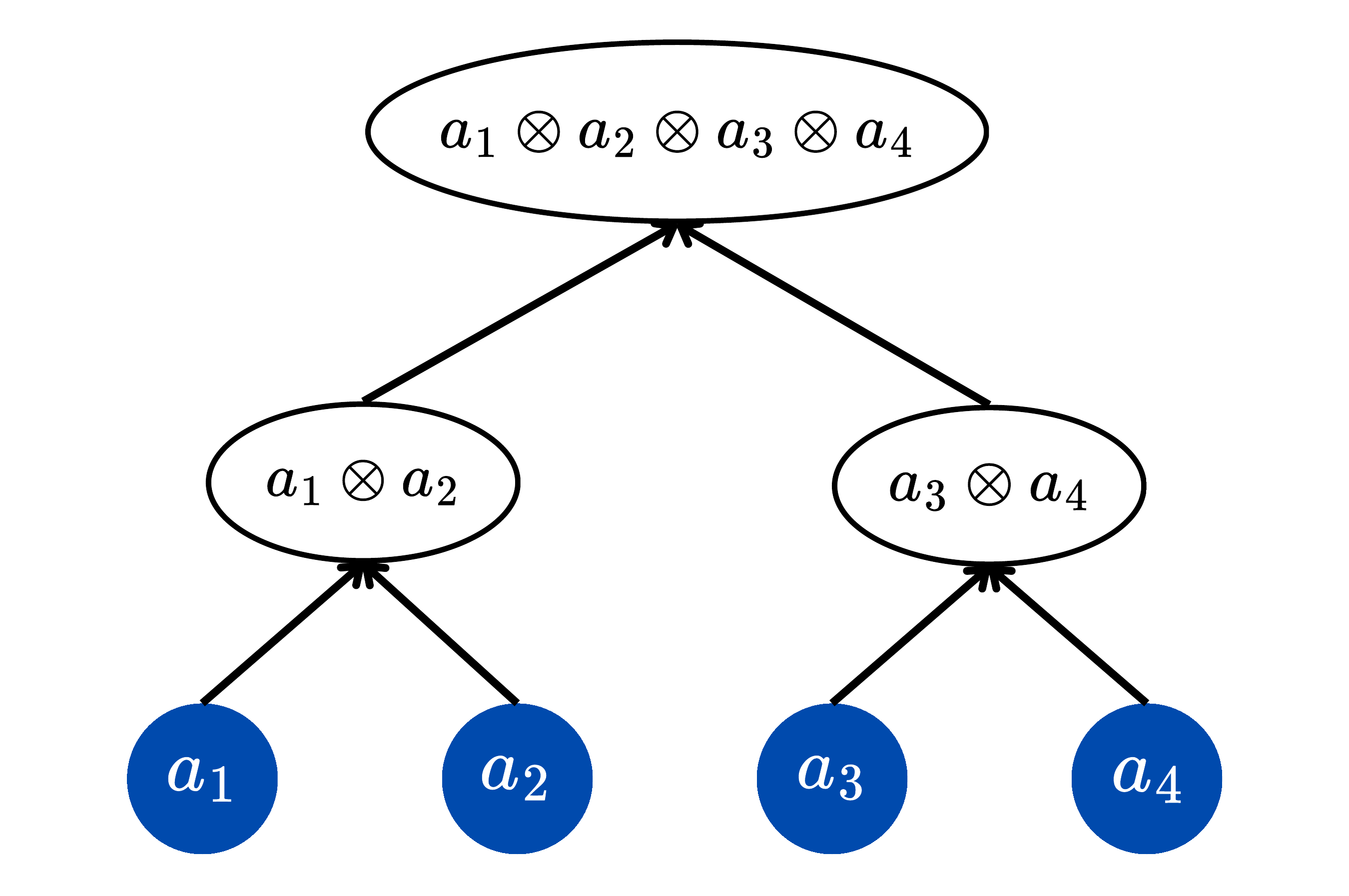}
    \caption{Upsweep phase of the parallel associative scan. Leaf nodes correspond to the ordered elements in $\mathcal{S}$ and parent nodes store partial reductgions computed bottom-up using $\otimes$.}
    \label{fig:upsweep}
\end{figure}

Due to the element's associative operator, combinations between two child nodes can be done in parallel. The depth of the binary tree is $\log_2 (n)$, resulting in a complexity of $\mathcal{O}(\log n)$ for the upsweep. 

The second phase of the algorithm is to perform a downsweep, recovering all terms in \eqref{eq:prefix_sums} using the partial reductions computed during the upsweep. We define the prefix with index $i$ as the reduction of all elements preceding $a_i$ in the ordered set $S$,
\begin{equation}
    p_i \coloneqq a_1 \otimes a_2 \otimes \cdots \otimes a_{i-1}, 
\end{equation}
with the convention that $p_1 = e$, where $e \in \mathcal{S}$ denotes the identity element of the associative operator. The identity element satisfies the property $e \otimes x = x \quad \forall x \in \mathcal{S}$.

During the downsweep, prefixes are propagated top-down along the same binary tree used in the upsweep. For a given internal node, the prefix represents the aggregation ($\otimes$) of all elements strictly to the left of that node's subtree. At the root of the tree, the prefix is initialized to the identity element $e$, since no elements precede the root's subtree. The prefix is then passed unchanged to the left child, since its subtree already captures all elements that precede it in the ordering. For the right child, the prefix is propagated by combining the parent's prefix with the left child's upsweep value, since the left subtree contains all elements that preceede the elements in the right subtree. In other words, the downsweep can be concisely defined as the following update rules propogated down the tree,
\begin{equation} \label{eq:downsweep_rules}
    \begin{aligned}
        & \text{pref(root)} = e \\
        & \text{pref}(L) = \text{pref}(P) \\
        & \text{pref}(R) = \text{pref}(P) \otimes \text{val}(L) \\
    \end{aligned}
\end{equation}
where for a given parent node $P$, it has left and right children, $L$, $R$, respectively. $\text{pref}(\cdot)$ denotes the prefix of the node and $\text{val}(\cdot)$ denotes the upsweeped value of the node. 
\begin{figure}[h]
    \centering
    \includegraphics[width=\columnwidth,
    trim=0 250pt 270pt 0,
        clip
    ]{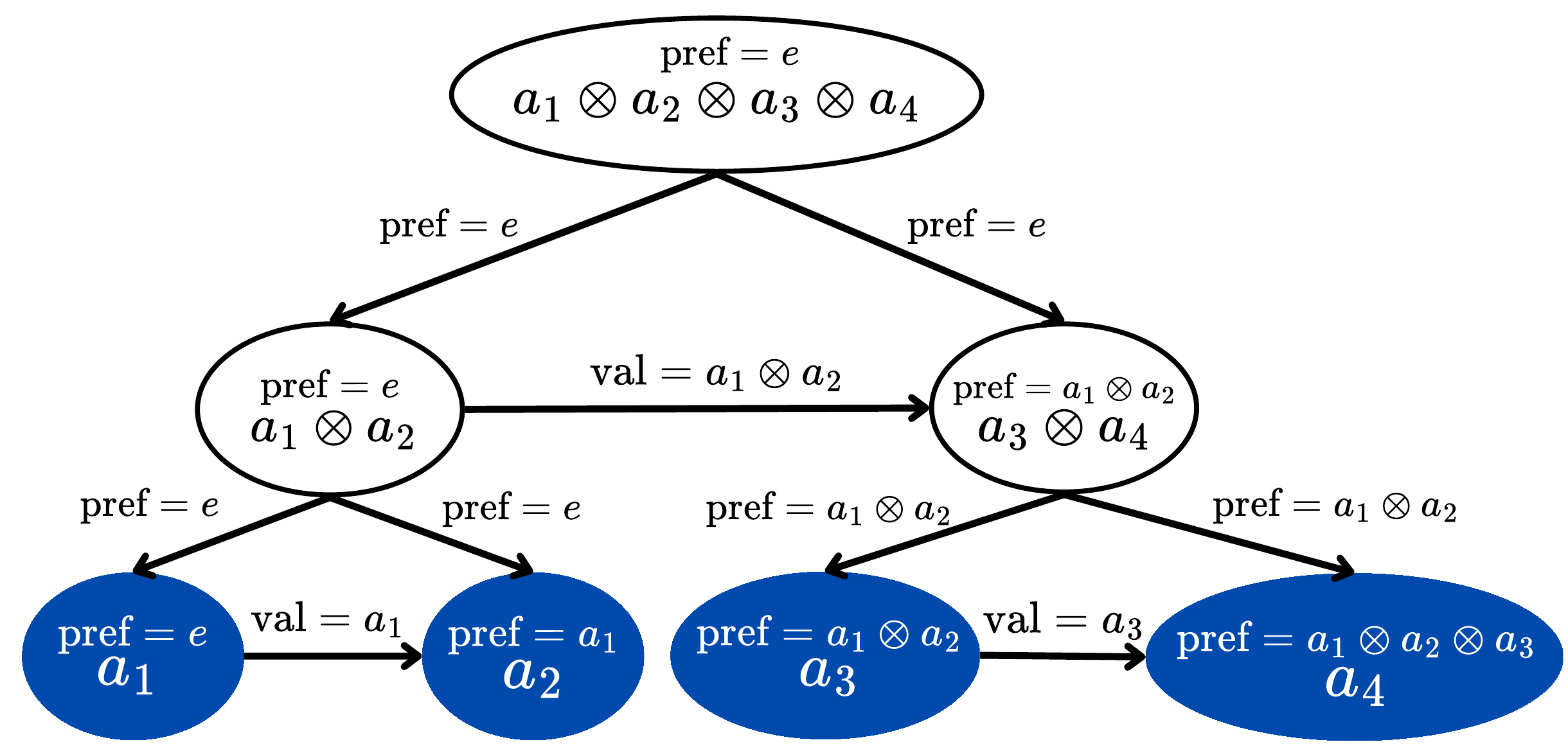}
    \caption{Downsweep phase of the parallel associative scan. Using the subtree aggregations during the upsweep, prefixes are propagated top-down along the same binary tree following the rules \eqref{eq:downsweep_rules}.}
    \label{fig:upsweep}
\end{figure}
For a given level, the prefix of each node can be calculated independently in parallel. Therefore, the complexity of the downsweep is $\mathcal{O}(\log n)$.

Lastly, each leaf node value can be combined with its associated prefix to recover elements \eqref{eq:prefix_sums}. Since both the upsweep and downsweep are $\mathcal{O}(\log n)$ complexity, parallel associative scans on parallel hardware have an overall complexity of $\mathcal{O}(\log n)$.

\section{FastSLS Duals $\tau$}\label{app:fastsls_duals}
This appendix briefly covers the purpose of the dual variables $\tau$ in Sec. \eqref{eq:sls_cost_matrices_tau} and how they are computed. As discussed in \ref{sec:sls}, FastSLS decomposes the robust policy optimization into two alternating procedures: first, an optimization of the tightened nominal trajectory, second, an optimization of a robust policy around that trajectory.

Though the two optimizations are performed separately, it is crucial for the controller update to capture which constraints are currently tight along the nominal trajectory. In doing so, this information allows the controller synthesis step to selectively penalize disturbance propagation at time steps where the constraints are active. Formally, we define the auxiliary terms $\beta_{k,j} \in \mathbb{R}^{n_c}$ and $\beta_{N,j} \in \mathbb{R}^{n_f}$,
\begin{equation}
    \begin{aligned}
        \beta_{k,j} &=
        \big\Vert
        \big(C^{(s)}_{k}\big)\mathbf{\Phi}^{\mathrm{x}}_{k,j}
        +
        \big(D^{(s)}_{k}\big)\mathbf{\Phi}^{\mathrm{u}}_{k,j}
        \big\Vert_{2,\mathrm{row}} ^2 \\
        & \forall j \in [N] \qquad \forall k \in [j, N]  \\
        \beta_{N,j} &= \| \left(C_N^{(s)}\right) \mathbf{\Phi}^{\mathrm{x}}_{N,j} \|^2_{2, \text{row}} \\
        & \forall j \in [N]
    \end{aligned}
\end{equation}
where $C^{(s)}_{k}, D^{(s)}_{k}, C^{(s)}_{N}$ are the linearized constraints \eqref{eq:lin_constraints_stage}-\eqref{eq:lin_constraints_term}, $\mathbf{\Phi}^{\mathrm{x}}_{k,j}, \mathbf{\Phi}^{\mathrm{u}}_{k,j}$ the system-level response matrices, and the square is applied elementwise. 

Using the unscaled dual variable $\lambda_k$ from \eqref{eq:al_admm}, we can calculate the duals $\tau$ by
\begin{equation}
    \tau_{k,j} = \frac{\lambda_k}{\sqrt{\beta_{k,j} + \epsilon}} \qquad \forall j \in [N] \quad  \forall k \in [j, N]
\end{equation}
where $\epsilon$ is some small number to circumvent points of non-differentiability. Intuitively, $\tau$ tells the controller step which constraints are the most costly to violate under disturbances, encouraging the controller to minimize the effect of the disturbance at that constraint.

\section{Dubins car}\label{app:dubins car}
We consider the planar Dubins car as a canonical nonholonomic system with bounded curvature. The system state at a discrete time step $k$ is
\begin{equation}
    x_k \coloneqq \begin{bmatrix}
        p_{x,k} \\
        p_{y,k} \\
        \theta_k
    \end{bmatrix}
\end{equation}
where $(p_{x,k}, p_{y,k})$ denote the planar position and $\theta_k$ the heading angle at a specific time step $k$. The control input is the angular velocity of the car
\begin{equation}
    u_k = \omega_k
\end{equation}
The dynamics are given by
\begin{equation}
        x_{k+1} = f(x_k, u_k) = \begin{bmatrix}
            p_{x,k} +  v \cos \theta_k\Delta t \\
            p_{y,k} + v \sin \theta_k\Delta t \\
            \theta_{k} + \omega_k\Delta t
        \end{bmatrix}
\end{equation}
where $v$ is a constant velocity and $\Delta t$ is the discrete time step.

We impose box constraints on the angular velocity
\begin{equation}
    \omega_{\text{min}} \leq w_k \leq \omega_{\text{max}}
\end{equation}
and enforce obstacle avoidance through state constraints on the position. For a given obstacle centered at $(c_x, c_y)$ with radius $r$, the corresponding constraint is defined as
\begin{equation} \label{eq:obstacle_constraints}
    r^2 - (p_x - c_x)^2 - (p_y-c_y)^2 \leq 0
\end{equation}
We consider a constant disturbance scaling matrix defined as $E = 2.5 \cdot 10^{-2} I_{3}$.

\section{Planar Quadrotor} \label{app:planar_quadrotor}
We consider a planar quadrotor with the following dynamics
\begin{subequations}
\begin{align}
\dot{x} &=
\begin{bmatrix}
v_x \\
v_y \\
\dot{\phi} \\
-\frac{1}{m}(u_1 + u_2)\sin(\phi) \\
\frac{1}{m}(u_1 + u_2)\cos(\phi) - g \\
\frac{L}{J}(u_2 - u_1)
\end{bmatrix}
\end{align}
\end{subequations}
with the state defined as $x := (p_x, p_y, \phi, v_x, v_y, \dot{\phi}) \in \mathbb{R}^6$ and input $u := (u_1, u_2) \in \mathbb{R}^2$. $(p_x, p_y)$ denotes the position, $\phi$ the pitch angle, $(v_x, v_y)$ defines the translational velocities, and $\dot{\phi}$ is the angular velocity. The inputs $(u_1, u_2)$ correspond to the individual rotor thrusts. We define the mass $m = 2.0576$, gravitational acceleration $g = 9.81$, arm length $L = 0.25$ and moment of inertia $J = 0.01$. We use a constant disturbance scaling matrix $E = 5 \cdot 10^{-2} \cdot \text{diag}(0, 0, 0, 1, 1, 0)$ and obstacles are defined similarly as in \eqref{eq:obstacle_constraints}.

\section{Quadruped and Humanoid} \label{app:quad_and_human}
In this appendix section, we briefly go over the environment of our legged simulation experiments, list their respective state space, control space, and dynamics function.

Our simulation environments for the quadruped and humanoid are derived from a modified version of the \textbf{mpx} environment from \cite{amatucci2025primal}. All simulation experiments are conducted in MuJoCo with all controls realized through the simulator's physics engine. Obstacle constraints are defined similarly as in \eqref{eq:obstacle_constraints}.

We use the Unitree Go2 EDU quadruped in both simulation and hardware. The quadruped contains 12 actuated joints: 4 hip joints, 4 thigh joints, and 4 knee joints. The state at time step $k$ is $x_k \in \mathbb{R}^{61}$, with the following structure
\begin{align*}
    x_{0:2}&: \text{Center of Mass (CoM) position} \\
    x_{3:6}&: \text{Floating base orientation (quaternion)} \\
    x_{7:18}&: \text{joint angles} \\
    x_{19:21}&: \text{CoM linear velocity } \\
    x_{22:24}&: \text{Floating base angular velocity} \\
    x_{25:36}&: \text{joint angular velocity} \\
    x_{37:48}&: \text{contact positions (world frame)} \\
     x_{49:60}&: \text{ground reaction forces at each feet}.
\end{align*}
The control input consists of joint torques, $u_k \in \mathbb{R}^{12}$ with each torque corresponding to one of the 12 joints of the quadruped. For our hardware experiments, we used a constant disturbance scaling matrix of $E = 2 \cdot 10^{-1} \cdot \text{diag}(1, 1, 0, 0, \ldots, 0)$ to capture uncertainty in the Vicon position estimation and sim-to-real model error. While this scaling matrix is hand-designed, we note that $E$ can be calibrated to yield high-probability guarantees on covering the true model error by following the approach of \cite{srinivasan2026safety}.

For the humanoid, we use a Unitree H1 model in our simulation experiments, which contains 19 actuated joints: 6 hip joints, 2 knee joints, 4 ankle joints, 4 shoulder joints, 2 elbow joints, and 1 torso joint. The state at time step $k$ is $x_k \in \mathbb{R}^{75}$, with the following structure
\begin{align*}
    x_{0:2}&: \text{CoM position} \\
    x_{3:6}&: \text{Floating base orientation (quaternion)} \\
    x_{7:25}&: \text{joint angles} \\
    x_{26:28}&: \text{CoM linear velocity } \\
    x_{29:31}&: \text{Floating base angular velocity} \\
    x_{32:50}&: \text{joint angular velocity} \\
    x_{51:62}&: \text{contact positions (world frame)} \\
    x_{63:74}&: \text{ground reaction forces at each feet}.
\end{align*}
The control input consists of joint torques, $u_k \in \mathbb{R}^{19}$ with each torque corresponding to one of the 19 joints of the humanoid. In our experiments. we define a constant disturbance matrix
\begin{subequations}
    \begin{align} \label{eq:humanoid_disturbance_def}
        E &= \operatorname{diag}(e_1, \dots, e_n), \quad \text{where} \\
        e_i &=
        \begin{cases}
        0.025, & i \in \{0,1,2\}, \\
        0.5,   & i \in \{26,27,28\}, \\
        0,     & \text{otherwise},
        \end{cases}
    \end{align}
\end{subequations}
defining disturbances in the humanoid's position and velocity. We run the algorithm for a maximum of 100 SQP iterations.

The system dynamics for both the quadruped and humanoid are provided by MuJoCo's rigid-body simulator, while ground reaction forces are computed explicitly according to the contact schedule. At each MPC planning step, a reference trajectory generator is used to define the locomotion pattern, which includes defining the gait sequence, swing and stance phases for each leg, and the foot placements over the MPC horizon.

\begin{figure}[t]
  \centering
  \includegraphics[width=\linewidth]{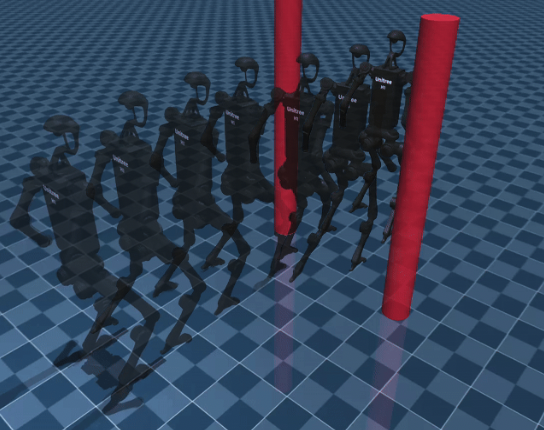} \vspace{-20pt}
  \caption{Whole-body humanoid (75D, 19C) robustly navigating through two obstacles using GPU-SLS in a Real-Time Iteration scheme. Our formulation is able to solve for robust control policies on average in \textbf{22 ms} on a NVIDIA RTX 4090 GPU, demonstrating the real-time capabilities of our formulation.} 
  \label{fig:humanoid_sls_navigation}
\end{figure}

\begin{figure}[t]
  \centering
  \includegraphics[width=\linewidth]{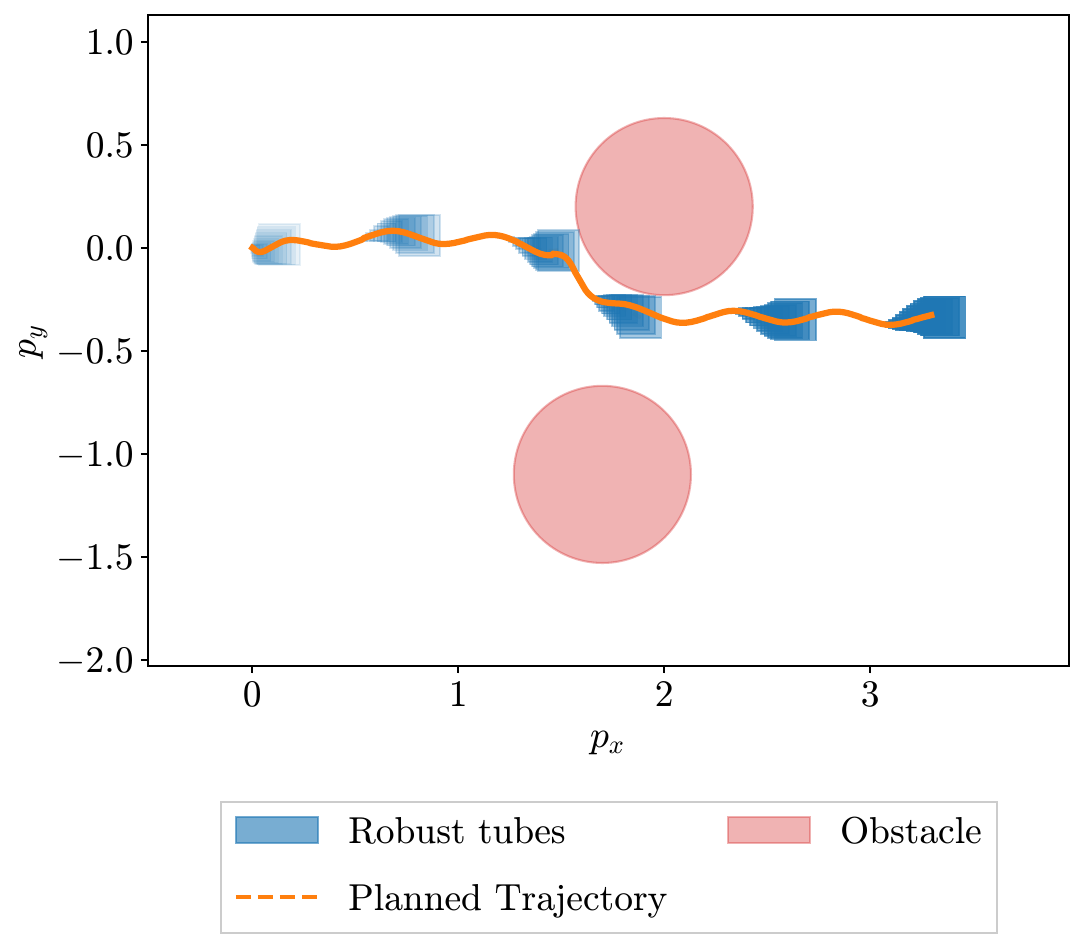} 
  \caption{Visualization of $(p_x, p_y)$-slices of the robust reachable tubes at certain time steps for the whole-body humanoid rollout in Fig.~\ref{fig:humanoid_sls_navigation}. At each time step, the calculated robust policy guarantees that our system stays within the calculated tubes, ensuring that the constraints are never violated even under external disturbances. Each obstacle is inflated to account for the size of the humanoid's torso. } 
  \label{fig:humanoid_overhead_tubes} \vspace{-15pt}
\end{figure}

\section{Humanoid RTI} \label{app:humanoid_rti_ext}
In this appendix section, we expand our experiments of RTI GPU-SLS to a whole rigid-body humanoid in simulation. We task a Unitree H1 humanoid robot with robustly navigating between two obstacles under a constant disturbance of $E = 2 \cdot 10^{-1} \cdot \text{diag}(1, 1, 0, 0, \ldots, 0)$. In Fig.~\ref{fig:humanoid_sls_navigation}, we show the humanoid safely navigating through the obstacles, and in Fig.~\ref{fig:humanoid_overhead_tubes} we show an overhead view of the obstacles and robust tube snapshots along the trajectory. To account for the robot's physical footprint, we inflate the obstacles by 0.33 meters. Across 700 RTI MPC steps, GPU-SLS calculates robust policies on average in 22 ms, demonstrating the real-time capabilities of our solver for different legged robots.

\section{Humanoid State Tubes} \label{app:humanoid_tubes_all_of_thm}
\begin{figure*}[p]
  \centering
  \includegraphics[
    width=\textwidth,
    height=\textheight,
    keepaspectratio
  ]{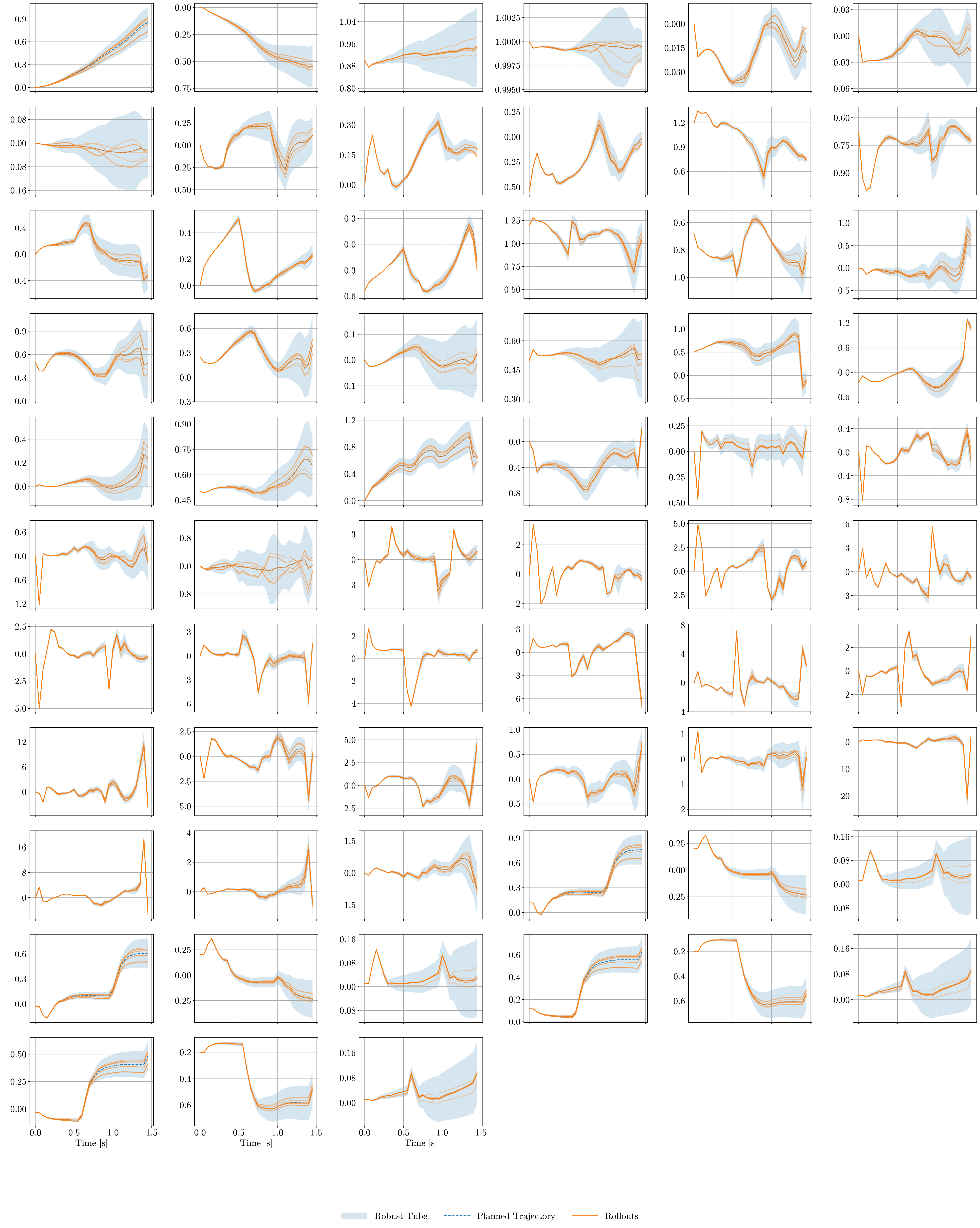}
  \caption{Simulated rollouts of Fig. \ref{fig:humanoid_timelapse}, showing disturbed states and robust tubes under adversarial disturbances. 
  }
  \label{fig:humanoid_tubes_all}
\end{figure*}

In this appendix section, we display the remaining state tubes (Fig.~\ref{fig:humanoid_tubes_all}) for the system in Fig.~\ref{fig:humanoid_tubes}. As shown, the humanoid stays within the computed robust tubes for all states, demonstrating the robustness guarantees of SLS.

\end{document}